\documentclass{article}

\PassOptionsToPackage{numbers, compress}{natbib}
\usepackage[preprint]{neurips_2023}

\usepackage[table,xcdraw,dvipsnames,svgnames]{xcolor}
\usepackage[utf8]{inputenc} 
\usepackage[T1]{fontenc}    
\usepackage{hyperref}       
\usepackage{url}            
\usepackage{booktabs}       
\usepackage{amsfonts}       
\usepackage{nicefrac}       
\usepackage{microtype}      

\usepackage[colorinlistoftodos, textwidth=30mm, shadow, textsize=small]{todonotes}
\definecolor{babyblue}{rgb}{0.54, 0.81, 0.94}
\definecolor{citrine}{rgb}{0.89, 0.82, 0.04}
\definecolor{misocolor}{rgb}{0.16,0.27,0.86}
\definecolor{jbcolor}{rgb}{0.9,0.4,0.2}
\definecolor{bernacolor}{rgb}{0.9608,0.4863,0.00}
\definecolor{carlcolor}{rgb}{0.0,0.9863,0.30}
\definecolor{grey}{rgb}{0.3, 0.3, 0.3}

\definecolor{graphicbackground}{rgb}{0.96,0.96,0.8}
\definecolor{rouge1}{RGB}{226,0,38}  
\definecolor{orange1}{RGB}{243,154,38}  
\definecolor{jaune}{RGB}{254,205,27}  
\definecolor{blanc}{RGB}{255,255,255} 
\definecolor{rouge2}{RGB}{230,68,57}  
\definecolor{orange2}{RGB}{236,117,40}  
\definecolor{taupe}{RGB}{134,113,127} 
\definecolor{gris}{RGB}{91,94,111} 
\definecolor{bleu1}{RGB}{38,109,131} 
\definecolor{bleu2}{RGB}{28,50,114} 
\definecolor{vert1}{RGB}{133,146,66} 
\definecolor{vert3}{RGB}{20,200,66} 
\definecolor{vert2}{RGB}{157,193,7} 
\definecolor{darkyellow}{RGB}{233,165,0}  
\definecolor{lightgray}{rgb}{0.9,0.9,0.9}
\definecolor{darkgray}{rgb}{0.6,0.6,0.6}
\definecolor{babyblue}{rgb}{0.54, 0.81, 0.94}
\definecolor{citrine}{rgb}{0.89, 0.82, 0.04}
\definecolor{misogreen}{rgb}{0.25,0.6,0.0}
\definecolor{PalePurp}{rgb}{0.66,0.57,0.66}
\definecolor{todocolor}{rgb}{0.66,0.99,0.99}
\definecolor{pearOne}{HTML}{2C3E50}
\definecolor{pearTwo}{HTML}{A9CF54}
\definecolor{pearTwoT}{HTML}{C2895B}
\definecolor{pearThree}{HTML}{E74C3C}
\colorlet{titleTh}{pearOne}
\colorlet{bull}{pearTwo}
\definecolor{pearcomp}{HTML}{B97E29}
\definecolor{pearFour}{HTML}{588F27}
\definecolor{pearFith}{HTML}{ECF0F1}
\definecolor{pearDark}{HTML}{2980B9}
\definecolor{pearDarker}{HTML}{1D2DEC}

\hypersetup{
	colorlinks,
	citecolor=pearDark,
	linkcolor=pearThree,
breaklinks=true,
	urlcolor=pearDarker}

\let\originalleft\left
\let\originalright\right
\renewcommand{\left}{\mathopen{}\mathclose\bgroup\originalleft}
\renewcommand{\right}{\aftergroup\egroup\originalright}

\renewcommand{\epsilon}{\varepsilon}

\newcommand{\nothere}[1]{}

\usepackage{xspace}

\usepackage{colortbl}
\usepackage{wrapfig}
\usepackage{booktabs}
\usepackage{graphicx}
\usepackage{enumitem}
\usepackage{algorithm}
\usepackage{listings}
\usepackage{pifont}
\usepackage{caption}
\usepackage{subcaption}
\usepackage{multirow}

\definecolor{RowHighlight}{gray}{0.9}
\usepackage{hyperref}

\usepackage[colorinlistoftodos]{todonotes}

\usepackage{etoolbox}
\makeatletter
\AfterEndEnvironment{algorithm}{\let\@algcomment\relax}
\AtEndEnvironment{algorithm}{\kern2pt\hrule\relax\vskip3pt\@algcomment}
\let\@algcomment\relax
\newcommand\algcomment[1]{\def\@algcomment{\footnotesize#1}}
\renewcommand\fs@ruled{\def\@fs@cfont{\bfseries}\let\@fs@capt\floatc@ruled
  \def\@fs@pre{\hrule height.8pt depth0pt \kern2pt}%
  \def\@fs@post{}%
  \def\@fs@mid{\kern2pt\hrule\kern2pt}%
  \let\@fs@iftopcapt\iftrue}
\makeatother

\title{Graph Contrastive Learning Meets Graph Meta Learning: A Unified Method for Few-shot Node Tasks}

\author{
    Hao Liu$^{1}$~~~Jiarui Feng$^{1}$~~~Lecheng Kong$^{1}$~~~Dacheng Tao$^{3}$~~~Yixin Chen$^{1}$~~~Muhan Zhang$^{2}$\\
    \texttt{\{liuhao, feng.jiarui, jerry.kong, ychen25\}@wustl.edu,}\\
    \texttt{dacheng.tao@gmail.com,~muhan@pku.edu.cn}\\
    ${}^1$Washington University in St. Louis\\
    ${}^2$Institute for Artificial Intelligence, Peking University\\
    ${}^3$JD Explore Academy\\}

\begin{document}

\maketitle

\begin{abstract}

Graph Neural Networks (GNNs) have become popular in Graph Representation Learning (GRL). One fundamental application is few-shot node classification. Most existing methods follow the meta learning paradigm, showing the ability of fast generalization to few-shot tasks.
However, recent works indicate that graph contrastive learning combined with fine-tuning can significantly outperform meta learning methods. Despite the empirical success, there is limited understanding of the reasons behind it. In our study, we first identify two crucial advantages of contrastive learning compared to meta learning, including (1) the comprehensive utilization of graph nodes and (2) the power of graph augmentations. To integrate the strength of both contrastive learning and meta learning on the few-shot node classification tasks, we introduce a new paradigm—\textbf{Co}ntrastive Few-Shot Node C\textbf{la}ssification (\textbf{COLA}). Specifically, COLA employs graph augmentations to identify semantically similar nodes, which enables the construction of meta-tasks without the need for label information. Therefore, COLA can utilize all nodes to construct meta-tasks, further reducing the risk of overfitting. Through extensive experiments, we validate the essentiality of each component in our design and demonstrate that COLA achieves new state-of-the-art on all tasks.

\end{abstract}

\section{Introduction}

Graph Neural Networks (GNNs)~\cite{hamilton2017inductive,kipf2016semi} have emerged as the predominant encoders for Graph Representation Learning (GRL) in recent studies, with node classification being a crucial area of investigation. Most researches focus on examining GNNs in supervised or semi-supervised settings~\cite{velivckovic2017graph,wu2020comprehensive}, which rely on many annotated data. Nevertheless, acquiring high-quality labels is challenging in many scenarios, leading to growing interest in exploring few-shot transductive node classification (FSNC), where only few labeled samples are provided for each class.

The majority of current studies on FSNC~\cite{GPN,TENT,G-Meta,MetaHG,MetaTNE,zhou2019meta,RALE} follow the meta learning~\cite{finn2017model,snell2017prototypical} paradigm. Specifically, to tackle a few-shot problem with $N$ classes and $k$ samples per class, meta learning gains knowledge through multiple training episodes with $N$-way $k$-shot meta-tasks generated from training classes. Each meta-task consists of a support set and a query set, both sampled from nodes belonging to a fixed number ($N$) of classes. The objective is to develop an algorithm that can perform well on the query set by training on only few support samples. This procedure enables the model to learn the latent distribution of tasks and thus can be easily transferred to tasks with unseen classes.

Self-supervised learning (SSL) can also effectively handle downstream few-shot tasks outside graph learning domains like computer vision~\cite{moco,SimCLR,BYOL}. Such capability demonstrates the importance of transferable and discriminative representations~\cite{tian2020rethinking} in few-shot learning. Observing the success of SSL in other areas, a recent study~\cite{TLP} on few-shot node classification used pre-trained node embeddings learned from existing Graph Contrastive Learning (GCL) methods~\cite{SUGRL,MERIT} to train a linear classifier for few-shot tasks. Even without label information, its best results significantly outperform the previous state-of-the-art (SOTA) supervised meta learning approaches.

To understand the success behind contrastive learning (CL), we analyze and validate two critical factors contributing to contrastive learning's exceptional performance through extensive experiments. The first factor is the use of data augmentation. It helps the model to learn discriminative embeddings with minimal redundant information from the graph, which is essential for few-shot tasks. Secondly, CL methods explicitly incorporate node embeddings from validation/test classes in contrastive loss, reducing the likelihood of model overfitting. In contrast, meta learning relies on node labels to construct meta-tasks, thus can only use nodes from training classes, losing much graph information.

Hence, one natural question emerges: \textsl{Can we leverage the advantages of contrastive learning to enhance the current meta learning framework?} To address this question, we propose a new paradigm for few-shot node classification termed \textbf{Co}ntrastive Few-Shot Node C\textbf{la}ssification (\textbf{COLA}). Unlike original meta-tasks, which require nodes within the same class to construct support sets, COLA construct meta-tasks without labels.

Specifically, the selection of support and query sets is the core of $N$-way $k$-shot meta-tasks construction. We start by randomly sampling $N$ query nodes and regard them as $N$ different ways. The main challenge is how to find $k$ semantically similar samples to each query node, without label information. 
We first use GNNs as graph encoders to get node embeddings in three augmented graphs. 
Leveraging the idea that similar nodes should maintain semantic similarity in perturbed graphs, the nodes that have similar embeddings to the query node across different augmented graphs are selected to construct the support set. To train an effective graph encoder, we generate the query embeddings from one augmented graph using a trainable GNN and obtain the query embeddings from another augmented graph by a momentum GNN, whose weight is the moving average of the trainable GNN.  

Our framework has several advantages: (1) We utilize the invariant information among three augmented graphs to construct semantically correct meta-tasks without label information; (2) Data augmentation allows the generation of a diverse range of meta-tasks over episodes and helps the GNN encoder learn a discriminative data representation; (3) The construction of meta-tasks enables the utilization of all nodes in training, further incorporating more graph information; (4) We take a slowly updated encoder to create a more stable support set candidate pool, which is less prone to noise compared to a rapidly updated encoder. We conduct tests on six real-world datasets, examining the necessity of each framework component. Our results demonstrate that the proposed framework outperforms SOTA approaches, highlighting its effectiveness and potential for application.

\section{Notations and Preliminaries}

We first introduce some preliminary concepts and notations. In this work, we consider an undirected attributed graph $\mathcal{G}=(\mathcal{V}, \mathcal{E}, \mathbf{A}, X)$, where $\mathcal{V}=\{v_1,\cdots,v_{|\mathcal{V}|}\}$ is the set of nodes, $\mathcal{E} = \{e_1,\cdots,e_{|\mathcal{E}|}\}$ is the set of edges. The adjacency matrix $\mathbf{A} \in \{0, 1\}^{|\mathcal{V}| \times |\mathcal{V}|}$ describes the graph structure, with $\mathbf{A}_{ij}=1$ indicating an edge between nodes $v_i$ and $v_j$ and $\mathbf{A}_{ij}=0$ otherwise. The feature matrix $X \in \mathbb{R}^{|\mathcal{V}| \times d}$ contains the node features, where $\mathbf{x}_i \in \mathbb{R}^d$ represents the feature of node $v_i$ and $d$ is the feature dimension. In our work, we focus on the node classification problem, where each node $i$ has a label $y_i \in C$ and $C$ is the set of labels with $|C|$ different classes. 

\textbf{Few-shot Node Classification.} 
In node classification, nodes are usually divided into train, validation, and test sets, denoted as $X_{train}$, $X_{val}$, and $X_{test}$, respectively. However, unlike supervised node classification where the node labels of train/validation/test sets are sampled from the same label set $C$, the label of nodes in few-shot learning are sampled from non-overlapped label sets for train/validation/test set, denoted as $C_{train}$, $C_{val}$. and $C_{test}$. Further, it holds that $C_{train} \cap C_{test} = \emptyset$. Few-shot Learning typically deals with $N$-way $k$-shot tasks, where the objective is to classify nodes into one of $N$ distinct classes using only $k$ labeled samples per class.

\textbf{Meta Learning.} 
Meta learning~\cite{finn2017model,snell2017prototypical} tries to solve the few-shot problems by designing a novel training strategy. The overall process of meta learning can be divided into meta-train and meta-test phases. During meta-train phase, the model is trained to simulate the few-shot learning environment. It enables the model to quickly adapt to new few-shot tasks with limited labeled data during the meta-test phase. Specifically, at each training episode, meta learning constructs an $N$-way $k$-shot task using samples from the training set $X_{train}$. To form an $N$-way $k$-shot task, meta learning first randomly select a set $C_{meta}$ with $N$ classes from $C_{train}$ and then generate a \textbf{support set} $\mathcal{S}=\{(\mathbf{x}_i,y_i)|y_i \in C_{meta}, i=1,\cdots , N\times k\}$ and a \textbf{query set} $\mathcal{Q}=\{(\mathbf{x}_i,y_i)|y_i \in C_{meta}, i=1,\cdots , N\times q\} (\mathcal{S} \cap \mathcal{Q} = \emptyset)$ by sampling $k$ support and $q$ query samples from each class in $C_{meta}$, respectively. The objective is to train on the support set so that it can perform well on the query set. In meta-test phase, the $N$-way $k$-shot tasks are constructed with samples in $X_{test}$ in a similar way.

\section{Contrastive Few-Shot Node Classification (COLA)}

In this section, we first identify two critical components that contribute hugely to the success of contrastive learning on FSNC but lack in meta learning. Next, we introduce our new paradigm COLA, which leverages the strengths of both contrastive learning and meta learning. Our key idea is to construct meta-tasks without labels, where the invariant information among three augmented graphs is utilized to construct semantically correct meta-tasks. The supervised contrastive loss~\cite{khosla2020supervised} is taken to learn the meta-tasks.

\subsection{Analysis on Success of Contrastive Learning in Few-Shot Node Classification}
\label{sec:3.1}

Although most current works on transductive FSNC follow meta learning framework (details will be discussed in Section~\ref{sec:related_work}), a recent study TLP~\cite{TLP} highlights the effectiveness of graph contrastive learning combined with fine-tuning. The authors conducted experiments using various existing graph contrastive learning methods and fine-tuned a linear classifier on top of the learned representation, which resulted in significant performance improvements on few-shot node classification tasks compared to SOTA supervised meta learning methods. 

To explain the strong performance of contrastive learning, we start to analyze the difference between contrastive learning and meta learning. Both techniques strive to bring the embeddings of semantically similar nodes closer and separate embeddings of semantically dissimilar ones.  However, the definition of semantically similar is different in the two methods. Meta learning regards all node embeddings from the same class as similar and node embeddings from different classes as dissimilar. In contrast, self-supervised contrastive learning only considers the embeddings of the same node in different augmented graphs as similar. A direct advantage of this is that contrastive learning can explicitly utilize all node embeddings in a given graph without worrying about label leaking issues. However, meta learning can only rely on samples from the training set, which may increase the likelihood of overfitting to the training classes and limit the model's ability to transfer knowledge to test classes. Further, leveraging the graph augmentation technique is another difference between contrastive learning and meta learning, which is already known to be effective in learning discriminative representation~\cite{tian2020rethinking}. We conjecture the above two differences contribute most to the success of contrastive learning in FSNC. 

\begin{wrapfigure}{r}{0.4\textwidth}
\vspace{-5pt}
\setlength{\columnsep}{0pt}%
	\centering
	\includegraphics[width=2in, trim={0.5cm, 0, 0.5cm, 0.5cm}]{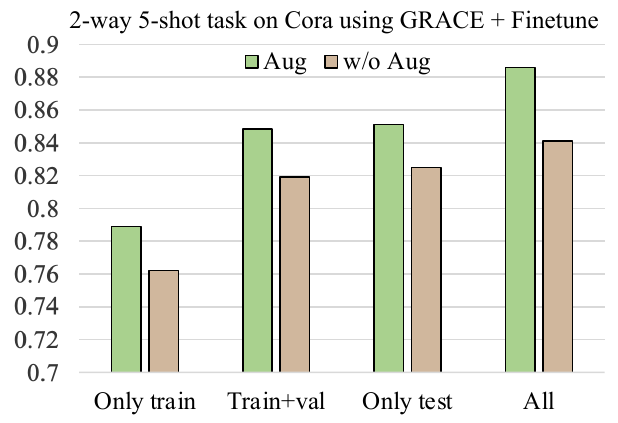}
 	\caption{2-way 5-shot task on Cora using GRACE+finetune. Accuracy of four situations w/ and w/o  augmentations.} 
    \label{fig:toy}
\vspace{-0pt}
\end{wrapfigure}

We then conduct extensive ablation studies to validate our speculations. We present one experimental result in Figure~\ref{fig:toy} and include other results in Appendix~\ref{app:exp}. The experiment is conducted on a 2-way 5-shot task from Cora~\cite{Cora-CiteSeer} dataset, and the node embeddings pre-trained from a GCL model named GRACE~\cite{GRACE} are used to train a classifier for few-shot tasks. We control the nodes used for pretraining to be sampled from $C_{train}$, $C_{train}\cup C_{val}$, $C_{test}$, and the whole graph. $C_{train}$, $C_{val}$ and $C_{test}$ contain 3, 2, 2 non-overlapped classes. We then assess the model on few-shot tasks sampled from $C_{test}$.

\begin{figure}
    
    \includegraphics[width=0.98\textwidth]{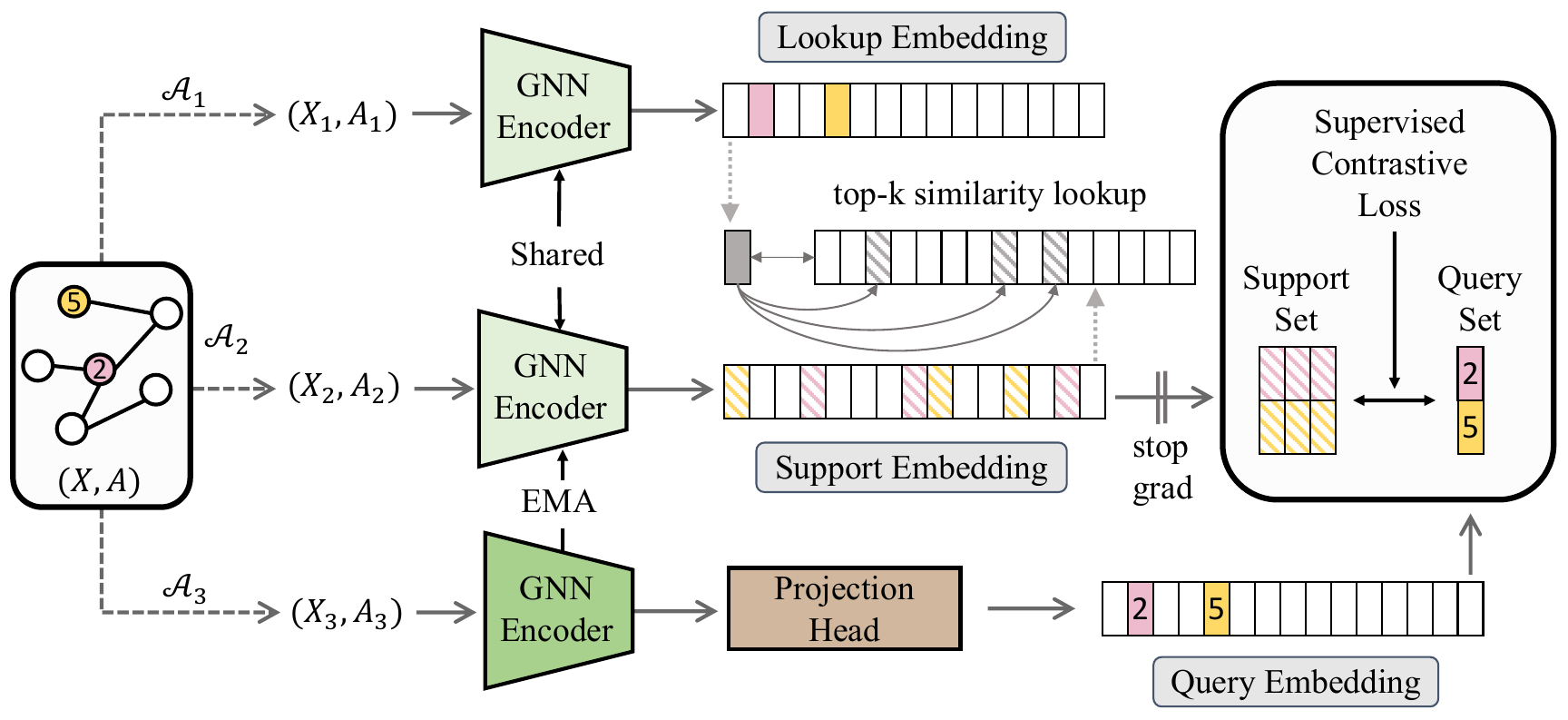}
    \caption{An overview of the COLA Framework. The construction of a 2-way 3-shot meta-task is illustrated. Two nodes 2 and 5 are sampled as the query set. The query node's embedding in Lookup Embedding matches with all node embeddings in Support Embedding. Top-$k$ similar embeddings are selected for the support set. Supervised contrastive loss is calculated for each task.}
    \label{fig:cofsnc}
\end{figure}

The results reveal several insights: although the number of nodes belonging to $C_{train}\cup C_{val}$ far exceeds the number of nodes in $C_{test}$, only using samples from $C_{test}$ to pretrain achieves better result than the other two settings. This experiment validates that explicitly leveraging test class samples during training can effectively avoid overfitting. Besides, using all nodes can maximize the utilization of graph information. Another observation is that eliminating augmentation leads to a performance decrease. Thus, the discriminative representation acquired by contrastive learning through data augmentation techniques is also crucial for few-shot tasks. 

From experimental results, we can see the explicit use of all nodes and data augmentation are indeed crucial to the performance of contrastive learning. These insights inspire us to propose a more robust meta learning framework that can effectively leverage the discriminative representation learned by contrastive learning while also benefiting from the generalization capabilities of meta learning. 

\subsection{Meta-task Construction without Labels}
In this section, we introduce our framework COLA, and the overall framework is illustrated in Figure~\ref{fig:cofsnc} and Algorithm~\ref{alg:code}. COLA aims to construct meta-tasks without labels, such that all nodes can be explicitly used during training. We will first introduce the process to generate three embeddings and explain how they function together to construct meta-tasks.

For a graph $\mathcal{G}$, let $\mathcal{A}(\mathcal{G})$ denotes the distribution of graph data augmentation of $\mathcal{G}$.  These augmentations \cite{GraphCL} typically involve one or more operations, such as node dropping, edge perturbation, and attribute masking. For the given graph represented as $(X,\mathbf{A})$, we apply three different data augmentations $\mathcal{A}_1, \mathcal{A}_2, \mathcal{A}_3 \sim \mathcal{A}$ and generate corresponding augmented graphs $(X_1, \mathbf{A}_1),(X_2, \mathbf{A}_2), (X_3, \mathbf{A}_3)$.

We then use GNNs to generate Lookup, Support, and Query Embeddings from the augmented graphs. Formally, 
\begin{equation}
    L:=f_{\textrm{ema}}(X_1, \mathbf{A}_1), \: S:=f_{\textrm{ema}}(X_2,\mathbf{A}_2),  \: Q:=g(f(X_3,\mathbf{A}_3)),
\end{equation}
where Lookup Embedding $L$ and Support Embedding $S$ are generated by a momentum encoder $f_\textrm{ema}$, and the Query Embedding $Q$ is generated by a trainable graph encoder $f$ with a projection head $g$. Weights of $f_{\textrm{ema}}$ are the moving average from $f$. Details about the momentum encoder will be discussed later. 


Then we present the process to construct meta-tasks. Inspired by contrastive learning, we first sample $N$ nodes and regard them as $N$ classes to form the query set $\mathcal{Q}$. Denote the query set $\mathcal{Q} = \{ v_1, \cdots, v_N \}$, where $v_i$ is the query node of the $i$-th way. To construct an $N$-way $k$-shot meta-task, the support set $\mathcal{S}$ should include $k$ samples that have similar semantics with the query sample from each of the $N$ ways. Then, how to find semantically similar samples is the main challenge. 

We first get query nodes' embeddings from Lookup Embedding $L$ and denote them as $\{ L_{v_1}, \cdots, L_{v_N} \}$. For each $i\in [1,\cdots, N]$, we then measure the similarity between $L_{v_i}$ and all node embeddings $\{  S_{1},\cdots, S_{|\mathcal{V}|} \}$ in Support Embedding $S$. The $k$ embeddings in $S$ with the highest similarity score will be selected as the support set, leading to $Nk$ samples in the support set. We denote them as $\{ S_{v_i^1}, \cdots, S_{v_i^k} \}_{i=1}^N$, where $S_{v_i^j}$ is the $j$-th support sample of the $i$-th query node. Finally, we get query nodes' embedding from Query Embedding $Q$ and denote them as $\{ Q_{v_1}, \cdots, Q_{v_N} \}$ and use them as the query set to construct a meta-task together with the support set. The task $\mathcal{T}$ can be represented as $\mathcal{T} = \{ Q_{v_i}, \{ S_{v_i^j} \}_{j=1}^k  \}_{i=1}^N$.


Actually, our method can be regarded as the process of matching and optimizing. We use the fact that the most essential graph information should be invariant across different augmented views. \textbf{Given a query node $v_i$, the $k$ embeddings (in $S$) that are most similar to $v_i$'s embedding from one augmented view should have comparable similarity to its embedding from another augmented view.} Consequently, after identifying the top $k$ embeddings most similar to the query node's lookup embedding $L_{v_i}$, we maximize the similarity between these $k$ embeddings and the query node's query embedding $Q_{v_i}$.


The momentum encoder is another component of meta-task construction. Formally, denote the parameters of $f_{\textrm{ema}}$ by $\theta_{\textrm{ema}}$ and parameters of $f$ by $\theta$, $\theta_{\textrm{ema}}$ is updated by exponential moving average (EMA) like \( \theta_{\textrm{ema}} = m \theta_{\textrm{ema}} + (1-m)\theta \), where $m$ is the momentum coefficient to control what degree it preserves the history. By employing a momentum encoder instead of the same trainable GNN encoder, the support set candidate pool ($S$) remains consistent across episodes and is less susceptible to noise or non-informative information from the rapidly changing encoder. Lookup Embedding and Support Embedding share the same momentum encoder, allowing for more accurate and consistent matches.

COLA is a new paradigm that constructs meta-tasks without labels. Including all nodes to meta-task construction effectively avoid overfitting to training classes. Unlike other graph meta learning methods, where support and query sets are derived from the original graph's embeddings, COLA constructs these sets from the embeddings of two distinct augmented views. By doing so, COLA can learn a discriminative representation. Since no label information is used, our framework uses exactly the same information as graph contrastive learning methods.
It's important to note that within this framework, the roles of the first and second augmented graphs can be interchanged. 
We perform extensive ablation studies to verify our designs and discuss the limitation in Appendix~\ref{app:limitation}.

\makeatletter
\lst@Key{spacestyle}{}
  {\def\lst@visiblespace{{#1\lst@ttfamily{\char32}\textvisiblespace{}}}}
\makeatother

\begin{algorithm}[t]
\caption{Pseudocode of COLA meta-task construction in a PyTorch-like style.}
\label{alg:code}
\definecolor{codeblue}{rgb}{0.25,0.5,0.5}
\lstset{
  backgroundcolor=\color{white},
  basicstyle=\fontsize{8pt}{8pt}\ttfamily\selectfont,
  columns=fullflexible,
  breaklines=true,
  captionpos=b,
  commentstyle=\fontsize{8pt}{8pt}\color{codeblue},
  keywordstyle=\fontsize{8pt}{8pt},
  showspaces=true,
  showstringspaces=false,
  spacestyle   = \color{white},
}
\begin{lstlisting}[language=python]
# f: GNN encoder, g: projection head, f_ema: momentum GNN encoder
# X: feature matrix, A: adjacency matrix, |V|: number of nodes
# N: number of classes, k: number of samples in support set, d: embedding dimension

x_1, x_2, x_3 = aug1(X,A), aug2(X,A), aug3(X,A)                   # three augmented versions

for q_idx in loader:                         # randomly sample N nodes with index q_idx: N*1
    Q = g(f(x_3))                                                   # Query Embedding: |V|*d
    L, S = f_ema(x_1), f_ema(x_2)                     # Lookup and Support Embeddings: |V|*d
    L, S = L.detach(), S.detach()            # No gradients to Support and Lookup Embeddings
    
    l = L[q_idx]                                  # lookup embeddings (surrogate key) l: N*d    
    sim = einsum("nd,vd->nv", [l, S])        # cosine similarity between l and S, sim: N*|V|
    s = top_k(sim, S)          # select top-k similarity samples in S for each way, v: N*k*d

    q = Q[q_idx]                                                   # query embeddings q: N*d
    loss = loss_sup(q,s)                   # calculate supervised loss following Equation (2)
\end{lstlisting}
\vspace{-0pt}
\end{algorithm}




\subsection{Meta-Train with Supervised Contrastive Loss}

\textbf{Meta-Train Phase.} 
To train the model, we employ supervised contrastive loss~\cite{khosla2020supervised}. In our setting, for each way $i$, the query embedding is treated as the anchor sample. The support embeddings $\{ S_{v_i^1},\cdots, S_{v_i^k} \}$ are considered as positive samples, while support embeddings $\{ S_{v_{i'}^1}, \cdots, S_{v_{i'}^k} \}_{i'\neq i}$ from other ways are viewed as negative samples. Formally, the pseudo-supervised contrastive loss for each meta-task can be expressed as follows:
\begin{equation}
    L_{sup} (\{Q_{v_i}, \{S_{v_i^j}\}_{j=1}^k \}_{i=1}^N )=-\sum_{i=1}^N \frac{1}{k} \sum_{j=1}^k \log \frac{\exp(Q_{v_i} \cdot S_{v_i^j} /\tau)}{\sum_{\mathbf{v}\in S_{t}} \exp(Q_{v_i} \cdot \mathbf{v} / \tau)},
\end{equation}
where $Q_{v_i}$ is the query sample of the $i$-th way, and $S_{v_i^j}$ is the $j$-th support sample of $Q_{v_i}$. $S_{t}$ denotes all the support embeddings in the current meta-task and $\tau$ is the temperature parameter.
Finally, the loss function of each meta-train episode is the average loss of multiple meta-tasks.

\textbf{Meta-Test Phase.} 
During the meta-test phase, we discard the momentum encoder and retain the GNN encoder. Then a linear classifier is trained on top of the learned node embeddings from the GNN encoder. To elaborate, we initially select $N$ classes from $C_{test}$ and sample $k$ labeled nodes from each class. The embeddings of these samples then undergo supervised training to fit a linear classifier. In the final step, we evaluate the performance using $q$ nodes from each of the $N$ classes.

\section{Related Work}
\label{sec:related_work}
\textbf{Graph Few-shot Learning.}
While GNNs for node classification are generally semi-supervised~\cite{kipf2016semi}, considerable efforts were spent on removing the labeling dependency~\cite{sun2020multi,hamilton2017inductive,velivckovic2017graph}. However, they cannot handle unseen classes during the test phase. This inspired research on the few-shot node classification problem. The majority of research employs a meta learning paradigm. Meta-GNN~\cite{zhou2019meta} adapts the optimization-based meta learning method MAML~\cite{finn2017model} to graph data. GFL~\cite{yao2020graph} enables few-shot classification on unseen graphs with seen node classes. GPN~\cite{GPN} uses ProtoNet~\cite{snell2017prototypical}, a metric-based meta learning method, and refines prototypes with the weights learned by a GCN~\cite{kipf2016semi}. G-Meta~\cite{G-Meta} leverages subgraph information and achieves good performance on both transductive and inductive FSCC. RALE~\cite{RALE} assigns relative and absolute locations to each node within meta-tasks. TENT~\cite{TENT} applies node-level, class-level, and task-level adaptations in each task to mitigate task variance impact. Recently, TLP~\cite{TLP}, inspired by graph contrastive learning, trains a few-shot classifier using pre-trained node embeddings, thereby significantly enhancing the performance over existing meta learning approaches. Its success prompts us to delve further into the potential of contrastive learning.

\textbf{Graph Contrastive Learning.}
Contrastive Learning methods~\cite{SimCLR, moco,BYOL} have been adapted to the graph domain. DGI~\cite{DGI} learns node representations by maximizing mutual information (MI) between local and global graph features. GRACE~\cite{GRACE} maximizes node-level agreement between two corrupted views. MVGRL~\cite{MVGRL} maximizes the MI between node representations of one view and graph representations of another view. GraphCL~\cite{GraphCL} applies various data augmentation techniques to the graph and then employs a contrastive loss function to move the representations of augmented views of the same graph closer. MERIT~\cite{MERIT} leverages bootstrapping within a Siamese network and multi-scale graph contrastive learning to enhance node representation learning. SUGRL~\cite{SUGRL} employs node embeddings from MLP as anchors and takes advantage of structural and neighbor information to obtain two kinds of positive samples. Different from previous methods, SUGRL takes the combination of triplet loss instead of InfoNCE loss~\cite{InfoNCE}. BGRL~\cite{BGRL} extends the non-contrastive setting~\cite{BYOL} that does not need negative samples to graph problem. TLP integrates various graph contrastive learning methods. SUGRL consistently delivers superior performance on few-shot tasks.

\textbf{Few-shot Learning with Contrastive Learning.}
Recent works in computer vision show that meta learning and contrastive learning can benefit from each other. Some recent few-shot auxiliary learning works~\cite{gidaris2019boosting,su2020does,chen2021pareto} view few-shot learning as the main task and combine the few-shot loss with self-supervised auxiliary tasks. ~\citet{liu2021learning} employs supervised contrastive learning on meta-tasks, where support images and query images are processed with different data augmentations to construct hard samples. CPLAE~\cite{gao2021contrastive} represents support and query samples using concatenated embeddings of both the original and augmented versions. It then regards prototypes of support samples as the anchor samples in contrastive learning. PsCo~\cite{PsCO} uses a momentum network with a queue like MoCo~\cite{moco} to improve pseudo labeling in the unsupervised meta learning setting. MetaContrastive~\cite{ni2021close} proposes a meta learning framework to enhance contrastive learning by transforming contrastive learning setup to meta-tasks. \textbf{However, in the field of graph learning, there is no work that enhances meta learning with the advantages of contrastive learning, and it is challenging to tailor these previous methods from the image domain for the graph.}\textbf{Few-shot Learning with Contrastive Learning.}

\section{Experiment}
In this section, we demonstrate COLA outperforms all the baselines in each task and provide the ablation study to validate the significance of each model component. 

\subsection{Datasets, Setup, and Baselines}

\textbf{Datasets.} We conducted our experiments on six benchmark datasets: Cora~\cite{Cora-CiteSeer}, CiteSeer~\cite{Cora-CiteSeer}, Amazon-Computer~\cite{Coauthor-Amazon} (Computer), CoraFull~\cite{CoraFull}, Coauthor-CS~\cite{Coauthor-Amazon} (CS), and ogbn-arxiv~\cite{ogb}. In each run for the same dataset, the classes were randomly divided into three subsets: $C_{train}$, $C_{val}$, and $C_{test}$. The setting of the split ratio follows previous works~\cite{TLP} and a detailed description of these datasets is provided in Appendix~\ref{app:data}. 

\textbf{Implementation Details.} We utilized Graph Convolutional Networks~\cite{kipf2016semi} (GCNs) as the encoder, and a multi-layer perceptron (MLP) as the projection head. Our data augmentation combines edge and feature dropout. The number of training tasks for calculating the average loss function is set to 20. We report mean accuracy and the 95\% confidence interval of 20 runs for both COLA and baseline models for a fair comparison. All models were tested on a single NVIDIA A100 80GB GPU. The detailed setting of hyperparameters is reported in Appendix~\ref{app:repro}.

\textbf{Baselines.} We compared our model with two groups of baselines: meta learning and graph contrastive learning with finetuning (proposed by TLP~\cite{TLP}). For meta learning, we first evaluate two plain meta learning models without GNN~\cite{kipf2016semi} as backbone: MAML~\cite{finn2017model} and ProtoNet~\cite{snell2017prototypical}, then we evaluate several meta learning works for few-shot node classification: Meta-GNN~\cite{zhou2019meta}, GPN~\cite{GPN}, G-Meta~\cite{G-Meta}, and TENT~\cite{TENT}. For TLP methods, we adhered to the settings and evaluated different graph contrastive learning methods for both contrastive-GCL and noncontrastive-GCL. They are MVGRL~\cite{MVGRL}, GraphCL~\cite{GraphCL}, GRACE~\cite{GRACE}, MERIT~\cite{MERIT}, SUGRL~\cite{SUGRL}, and BGRL~\cite{BGRL}, respectively.

\subsection{Main Results}
\begin{table}[t]
\small
\caption{Results on Cora, CiteSeer and CoraFull datasets. (Top rows) Meta Learning. (Middle rows) Graph Contrastive Learning with fine-tuning. (Bottom row) COLA (our method). All scores are averaged over 20 runs. Evaluation metrics were scaled to 100 for readability purposes. In bold are methods with the best results for each task. In blue are methods with the best results in each group.}
\setlength{\tabcolsep}{4.4pt}
\renewcommand{\arraystretch}{1.15}
\begin{tabular}{@{}lllllll}
\toprule
Dataset     & \multicolumn{2}{c}{Cora}    & \multicolumn{2}{c}{CiteSeer} & \multicolumn{2}{c}{CoraFull} \\
\cmidrule(lr){2-3} \cmidrule(lr){4-5} \cmidrule(lr){6-7}
Task        & 2-way 1-shot & 2-way 5-shot & 2-way 1-shot  & 2-way 5-shot & 5-way 1-shot  & 5-way 5-shot \\
\midrule
\multicolumn{7}{c}{\cellcolor{pearDark!25}Meta learning}                                 \\
\midrule
MAML~\cite{finn2017model}                & 52.59 $\pm$  2.28   & 56.45 $\pm$  2.41   & 51.77 $\pm$  2.28    & 54.21 $\pm$  2.30   & 22.47 $\pm$  1.21    & 26.58 $\pm$  1.32   \\
ProtoNet~\cite{snell2017prototypical}    & 51.69 $\pm$  2.17   & 55.00 $\pm$  2.39   & 51.43 $\pm$  2.12    & 53.23 $\pm$  2.28   & 34.17 $\pm$  1.74    & 46.86 $\pm$  1.74   \\
Meta-GNN~\cite{zhou2019meta}             & 57.87 $\pm$  2.52   & 57.35 $\pm$  2.30   & 55.12 $\pm$  2.62    & 60.59 $\pm$  3.26   & 55.36 $\pm$  2.49    & 71.42 $\pm$  2.02   \\
GPN~\cite{GPN}                           & 56.09 $\pm$  2.08   & 63.83 $\pm$  2.86   & 59.33 $\pm$  2.23    & 65.60 $\pm$  2.47   & 56.48 $\pm$  2.72    & 71.23 $\pm$  2.11   \\
G-Meta~\cite{G-Meta}                     & \cellcolor{pearDark!10}66.15 $\pm$  3.00   & \cellcolor{pearDark!10}82.85 $\pm$  1.19   & 54.33 $\pm$  2.02    & 61.47 $\pm$  2.37   & \cellcolor{pearDark!10}58.47 $\pm$  2.37    &     \cellcolor{pearDark!10}72.03 $\pm$  1.88         \\
TENT~\cite{TENT}                         & 54.33 $\pm$  2.10   & 58.97 $\pm$  2.40   & \cellcolor{pearDark!10}60.06 $\pm$  3.01    & \cellcolor{pearDark!10}66.31 $\pm$  2.45   & 49.83 $\pm$  2.02    & 64.23 $\pm$  1.75   \\
\midrule
\multicolumn{7}{c}{\cellcolor{pearDark!25}Graph Contrastive Learning  +   Finetune}         \\
\midrule
BGRL~\cite{BGRL}         & 59.16 $\pm$  2.48   & 81.31 $\pm$  1.89   & 54.33 $\pm$  2.14    & 66.74 $\pm$  2.13   & 40.82 $\pm$  1.95    & 69.98 $\pm$  1.67  \\
MVGRL~\cite{MVGRL}       & 74.96 $\pm$  2.94   & 91.32 $\pm$  1.47   & 63.39 $\pm$  2.69    & 79.73 $\pm$  1.92   & 66.40 $\pm$  2.31    & 83.99 $\pm$ 1.51   \\
MERIT~\cite{MERIT}       & 70.63 $\pm$  3.11   & 91.00 $\pm$  1.22   & 65.64 $\pm$  2.94    & 78.54 $\pm$  2.43   & 65.17 $\pm$  1.96    & \cellcolor{pearDark!10}84.74 $\pm$ 1.44   \\
GraphCL~\cite{GraphCL}   & 74.32 $\pm$  3.26   & 90.43 $\pm$  1.21   & 71.39 $\pm$  3.17    & 79.60 $\pm$  1.89   & 66.76 $\pm$  2.75    & 84.55 $\pm$ 1.48    \\
GRACE~\cite{GRACE}       & 71.50 $\pm$  1.42   & 88.49 $\pm$  1.44   & 67.43 $\pm$  2.51    & 82.09 $\pm$  1.64   & 62.05 $\pm$  2.22    & 81.54 $\pm$  1.52   \\
SUGRL~\cite{SUGRL}       & \cellcolor{pearDark!10}81.52 $\pm$  2.09   & \cellcolor{pearDark!10}92.49 $\pm$  1.02   & \cellcolor{pearDark!10}72.43 $\pm$ 2.42     & \cellcolor{pearDark!10}86.58 $\pm$  1.19   & \cellcolor{pearDark!10}73.95 $\pm$  2.13    & 83.07 $\pm$  1.21   \\
\midrule
COLA (ours) & \textbf{84.58 $\pm$  1.96}   & \textbf{94.03 $\pm$  1.48}   & \textbf{76.54 $\pm$  2.02}    & \textbf{86.87 $\pm$  1.49}   &\textbf{74.36 $\pm$  2.37}    & \textbf{86.59 $\pm$  2.26}  \\
\bottomrule

\end{tabular}
\label{tabmain}
\end{table}

Evaluations were made under 2-way 1-shot/5-shot settings on Cora, CiteSeer, and Computer datasets due to the limited number of available classes. CoraFull, CS, and ogbn-arxiv datasets were evaluated under 5-way 1-shot/5-shot settings. We present the main results on Cora, CiteSeer, and CoraFull datasets in Table~\ref{tabmain} and include results on other datasets in Appendix~\ref{app:exp}.

\textbf{Our method COLA outperforms all the other baselines in every task.} Compared with meta learning methods, COLA achieves at least 11.18\% and up to 20.56\% absolute accuracy improvement. The results demonstrate that the utilization of all nodes and a discriminative data representation indeed benefit the learning on few-shot tasks. Thus even constructing meta-tasks without label information, COLA can achieve excellent performance over traditional meta learning methods.

Graph contrastive learning methods benefit from the learned discriminative representations and show excellent ability to deal with downstream few-shot tasks. SUGRL achieves the best performance on most few-shot tasks. COLA outperforms SUGRL in each task with a maximum relative accuracy improvement of 5.93\%. This demonstrates that the use of $N$-way $k$-shot task construction in COLA makes it more suitable for few-shot problems compared to contrastive learning methods.

\subsection{Model Design Component Analysis}

\subsubsection{Query, Support, Lookup Embeddings}
\label{sec:ablation5.3.1}
First, we examine the primary design elements of COLA: Query ($Q$), Support ($S$), and Lookup Embeddings ($L$) and present the results in Table~\ref{tabab1}. To understand the distinct function of each one, we investigate three alternative scenarios.
In the first scenario, we only use the Query Embedding. The query sample $Q_{v_i}$ is extracted from the Query Embedding and has to align with all nodes within the Query Embedding itself to identify the support set. The second scenario omits the Lookup Embedding. Here, query sample $Q_{v_i}$ is compared with all nodes from Support Embedding to find the top-$k$ similar ones in $S$. The third scenario excludes Support Embedding, so we use the query embedding $L_{v_i}$ from Lookup Embedding to compare with all node embeddings in Query Embedding.

Compared with COLA, the first and second scenarios regard the query embedding as its own lookup tool. This reduces the amount of information in the meta-task, leading to suboptimal results. In the second scenario, the use of Support Embedding further deteriorates the performance since the inconsistency between the Query and Support Embeddings' encoders leads to a mismatch. The third scenario involves the Lookup Embedding, but both the query and support set are derived from Query Embedding, which means the model cannot take advantage of the extra information gained from two different augmented views. We also find that even some of these suboptimal setups can still outperform meta learning methods, underscoring the importance of using all available nodes.

Our COLA model significantly outperforms the three scenarios, illustrating the importance of each component in its design. In essence, we benefit from the invariant information among these three augmented graphs to construct meta-tasks, such that the support set selected by Lookup Embedding has very similar semantics to the query set. This ensures our model keeps constructing semantically correct meta-tasks.

\begin{table}[t]
\small
\caption{Component Analysis of Query ($Q$), Support ($S$), Lookup ($L$) Embeddings on Cora and CiteSeer datasets. The first three rows control different components in meta-task construction. The last row is COLA's setting. In bold are the best results, and underlines are the second best ones. }
\setlength{\tabcolsep}{5pt}
\renewcommand{\arraystretch}{1.1}
\begin{tabular}{cccllllll}
\toprule
      &   &  & \multicolumn{3}{c}{Cora}    & \multicolumn{3}{c}{CiteSeer} \\
 \cmidrule(lr){4-6} \cmidrule(lr){7-9}
 $Q$& $S$& $L$& 2-way 1-shot  & 2-way 3-shot & 2-way 5-shot & 2-way 1-shot  & 2-way 3-shot & 2-way 5-shot \\
\midrule
\multicolumn{1}{c}{$\checkmark$} &                                                                     & &   61.90 $\pm$ 1.26          & 84.12 $\pm$ 2.24 & \underline{88.24 $\pm$ 1.89} & 56.03 $\pm$  1.73 & \underline{71.46 $\pm$ 2.97}& \underline{74.69 $\pm$ 2.22} \\
\multicolumn{1}{c}{$\checkmark$} & \multicolumn{1}{c}{$\checkmark$}                                    & & 75.79 $\pm$ 2.75 & 75.20 $\pm$ 2.68 & 79.44 $\pm$ 2.01 & 59.48 $\pm$ 2.83 & 63.73 $\pm$ 2.48 & 69.10 $\pm$ 2.31 \\
\multicolumn{1}{c}{$\checkmark$} && \multicolumn{1}{c}{$\checkmark$}                                     & \underline{76.24 $\pm$ 3.68} & \underline{86.47 $\pm$ 1.45} & 85.78 $\pm$ 2.57 & \underline{64.42  $\pm$ 2.34} & 69.33 $\pm$ 3.15 & 73.13 $\pm$ 2.27 \\

\multicolumn{1}{c}{\textcolor{pearDark!100}{$\checkmark$}} & \multicolumn{1}{c}{\textcolor{pearDark!100}{$\checkmark$}} & \multicolumn{1}{c}{\textcolor{pearDark!100}{$\checkmark$}} & \textbf{84.58 $\pm$  1.96}   & \textbf{92.29 $\pm$ 1.71}& \textbf{94.03 $\pm$  1.48}   & \textbf{76.54 $\pm$  2.02}   & \textbf{80.26 $\pm$  2.72} & \textbf{86.87 $\pm$  1.49}  \\
\bottomrule

\end{tabular}
\label{tabab1}
\vspace{-5pt}
\end{table}




\begin{wraptable}{r}{6.5cm}
\vspace{-13pt}
\small
\caption{Relationship between the momentum parameter and accuracy on CiteSeer.}
\setlength{\tabcolsep}{3pt}
\renewcommand{\arraystretch}{1.1}
\begin{tabular}{@{}llllll}
\toprule
momentum   & 0     & 0.5 & 0.8   & 0.9   & 1 \\ \midrule
2-way 1-shot & 70.46 & 74.47    & 75.13 & \cellcolor{pearDark!10}76.54 &  54.85 \\ 
2-way 5-shot & 78.05 & 81.09    & 83.34 & \cellcolor{pearDark!10}86.87 &  65.43 \\ \bottomrule
\end{tabular}
\label{tabmm}
\vspace{-5pt}
\end{wraptable}

\subsubsection{Momentum Encoder}

To generate Support and Lookup Embedding, COLA uses a momentum GNN encoder $f_{\textrm{ema}}$, whose weights are an exponential moving average of the weights of the trained GNN encoder. The momentum encoder is more stable than the trainable GNN encoder. We test different values of the momentum variable from 0 to 1 and present the results in Table~\ref{tabmm}. Value 0 means the encoder is updated to the trained GNN encoder at each step and value 1 means the encoder is never updated. The results show that using the shared weight encoder (momentum=0) will harm the model performance. A static encoder (momentum=1) always contains the exact same information and constrains the information support embeddings can bring. A larger momentum (around 0.9) can help the momentum encoder memorize historical information, contributing to a consistent and stable Support Embedding.

\begin{figure}
\centering
\begin{subfigure}{.32\textwidth}
    \centering
    \includegraphics[width=1\linewidth,trim={1cm, 0cm, 0cm, 2.5cm}]{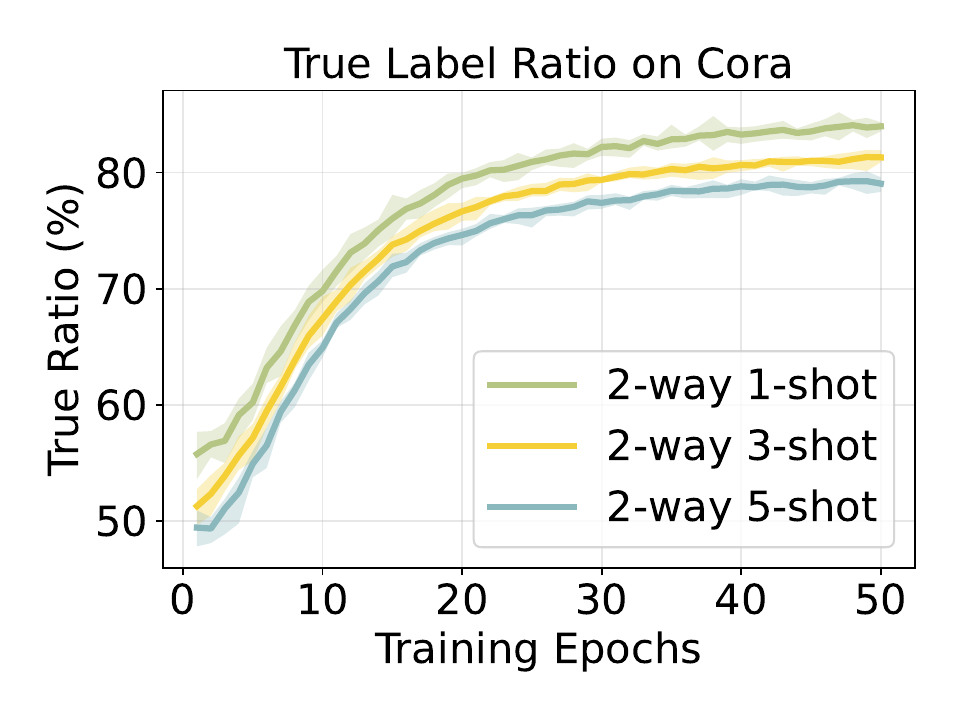}  
    \caption{True Label Ratio on Cora}
    \label{fig:tlcora}
\end{subfigure}
\begin{subfigure}{.32\textwidth}
    \centering
    \includegraphics[width=1\linewidth,trim={1cm, 0cm, 0cm, 2.5cm}]{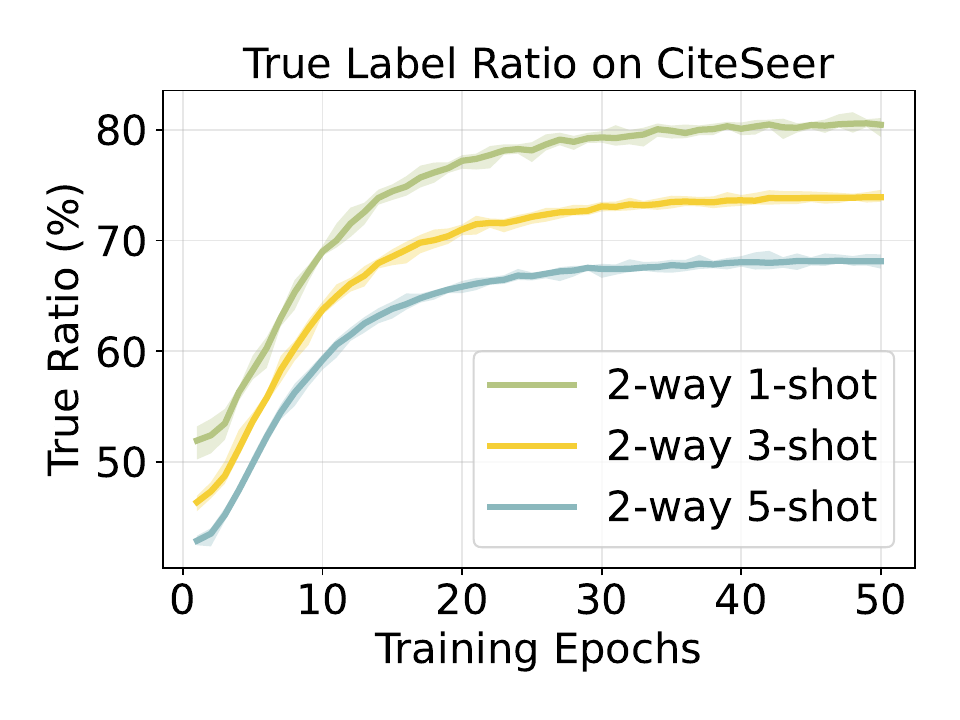}  
    \caption{True Label Ratio on CiteSeer}
    \label{fig:tlciteseer}
\end{subfigure}
\begin{subfigure}{.32\textwidth}
    \centering
    \includegraphics[width=1\linewidth,trim={1cm, 0cm, 0cm, 2.5cm}]{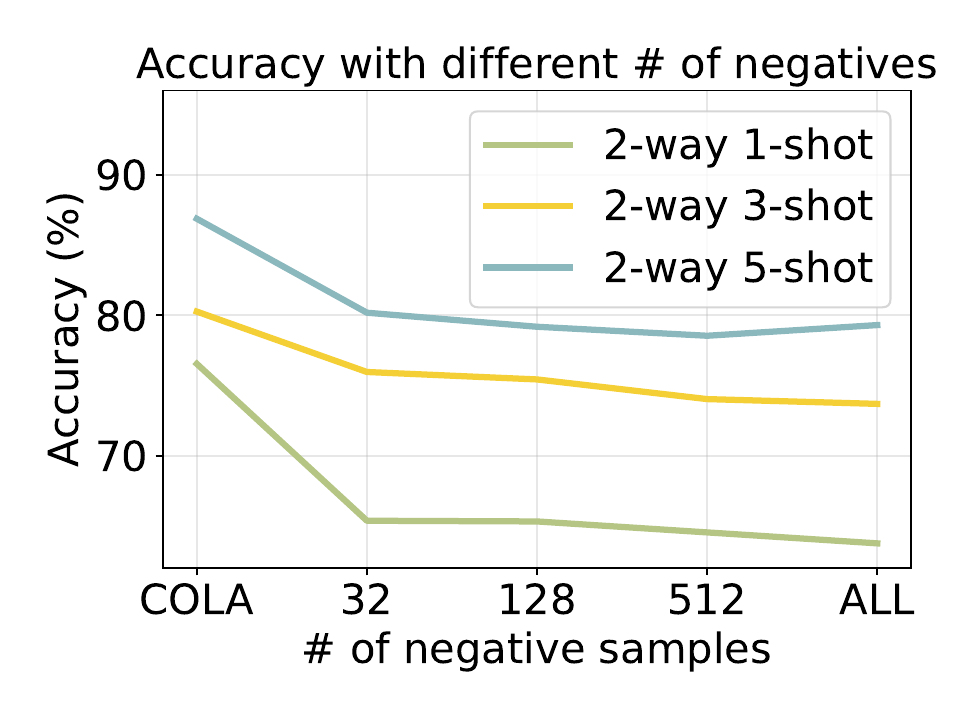}  
    \caption{Increasing the \# of negatives}
    \label{fig:neg}
\end{subfigure}
\caption{(a) and (b): true label ratio that measures the ratio of the selected support samples actually having the same label as the query sample. (c): Performance drops with extra negative samples.}
\label{FIGURE LABEL}
\end{figure}
\vspace{-5pt}

\subsection{Deep Investigation of COLA}
\textbf{True Label Ratio.} 
To evaluate the quality of task construction, we measure the true label ratio in each task. The true label ratio is calculated by $R_{true}={n}_{t}/{Nk}$, where ${n}_{t}$ is the number of selected support samples that indeed have the same label with corresponding query sample, and $Nk$ is the total number of support samples. To better visualize the trend, we only present the true label ratio within 50 epochs in Figure~\ref{fig:tlcora} and \ref{fig:tlciteseer}. Note that $R_{true}$ still increase after epoch 50.
The trend of $R_{true}$ reflects that the model is gradually selecting more and more support nodes that have exactly the same label as the query node.
For example, the initial true label ratio for Cora's 2-way 5-shot problem is around 0.41 and it steadily increases to 0.8, indicating that only around 2 selected support samples in this task have false labels. This measure verifies that the proposed method can construct semantically correct meta-tasks even without label information.

\textbf{Analysis of the number of negatives.}
Contrastive learning methods benefit from both the data augmentation and the large number of negative samples. In COLA, although we take the supervised contrastive loss, the number of negative samples is relatively small. This is because all the negative samples of a node only come from the support set of other ways, e.g. $(N-1)k$ for a $N$-way $k$-shot problem. Consequently, we examine whether the meta-tasks constructed by COLA will benefit from a large number of negative samples just like contrastive learning does. Thus, we vary the number of negatives from $(N-1)k$ to $|\mathcal{V}|$ (number of nodes in the graph) and present the result in Figure~\ref{fig:neg}. We get a conclusion that is contrary to expectations: the performance of our model is negatively impacted by increasing the number of negative samples in each case. 
We conjecture that the advantages contrastive learning gains from a high number of negative samples do not transfer well to few-shot tasks. Consequently, it underscores the need for a unified method (COLA) that is more suitable for FSNC.

\textbf{Using all nodes and data augmentation indeed contributes to the success of COLA.} 
We evaluate whether the utilization of all nodes and data augmentation will be helpful in our model and show the 
results of the CiteSeer dataset in Table~\ref{tabaug}. 
From the results, we can conclude that training without 
\begin{wraptable}[8]{r}{5cm}
\vspace{-10pt}
\small
\caption{Results of w/ and w/o augmentations and nodes from $C_{test}$.}
\setlength{\tabcolsep}{3pt}
\renewcommand{\arraystretch}{1.1}
\begin{tabular}{@{}lccc}
\toprule
  & $C \textbackslash C_{test}$       & $C_{test}$   & All nodes \\ \midrule
w/ aug &  68.43    & 72.19 &  \cellcolor{pearDark!10}86.97 \\ 
w/o aug & 65.18    & 61.02 &  74.51 \\ \bottomrule
\end{tabular}
\label{tabaug}
\vspace{-5pt}
\end{wraptable}
all nodes will lead to a performance decrease, especially when nodes from $C_{test}$ are not involved. 
Data augmentation is also important for our method since the meta-task construction relies on invariant graph information across the three augmented views.
These findings underscore the fact that COLA significantly benefits from data augmentation, enabling the construction of meta-tasks that optimally leverage graph information.



\section{Conclusion}

In this paper, we focus on the transductive few-shot node classification. We first identify several key components behind the success of contrastive learning on FSNC, including the comprehensive use of graph nodes and the power of graph augmentations.
We then introduce a new paradigm—\textbf{Co}ntrastive Few-Shot Node C\textbf{la}ssification (\textbf{COLA}).  
Unlike traditional meta learning methods that require label information, COLA finds semantically similar node embeddings to construct meta-task by leveraging the invariant information across three augmented graphs. 
COLA contains the advantages of both contrastive learning and meta learning on the few-shot node classification tasks. Through extensive experiments, we validate the essentiality of each component in our design and demonstrate that COLA achieves new state-of-the-art on all tasks. One limitation of our method is the increased computational cost due to the sorting operation used to find the support set, though this increase is linear to $|\mathcal{V}|$ and has no significant negative impacts in practice.
We believe our research will bring some new insights to the FSNC field.

\bibliographystyle{unsrtnat}
\bibliography{ref}

\newpage
\appendix
\section{Dataset Description}
\label{app:data}

We provide the dataset description in this section. The statistic descriptions are provided in Table ~\ref{tab:dataset}. Note that two classes in CoraFull dataset have extreme few samples, and cannot provide enough query set samples for validation or test, thus these two classes are omitted in all experiments. 

\textbf{Cora}~\cite{Cora-CiteSeer} is a citation network comprised of scientific publications that are categorized into one of seven classes. Here, each node represents a publication, with an edge indicating a citation relationship between two nodes. Each node features a binary word vector signifying the absence or presence of a word from a predefined dictionary.

\textbf{CiteSeer}~\cite{Cora-CiteSeer} is also a citation network similar to Cora, where nodes represent scientific documents and edges represent citations.

\textbf{Amazon-Computer}~\cite{Coauthor-Amazon} is a product co-purchasing network where nodes represent products and edges represent that the two products are frequently bought together. The label of each node is the product category it belongs to.

\textbf{Coauthor-CS}~\cite{Coauthor-Amazon} is a collaboration network, where nodes represent authors and edges indicate the collaboration relationship between two authors. The node feature includes information about publishing history, affiliations, and research interests. The label represents the field of the author's research.

\textbf{CoraFull}~\cite{CoraFull} is also a citation network, which is extracted from the original data from the entire Cora network. Cora is a small subset of CoraFull.

\textbf{Ogbn-arxiv}~\cite{ogb} is a part of Open Graph Benchmark collection. It consists of the co-authorship derived from ArXiv, where nodes represent academic preprint papers and edges represent whether the two papers share at least one same author.

\begin{figure}[h]
\centering
\begin{subfigure}{.32\textwidth}
    \centering
    \includegraphics[width=1\linewidth,trim={1cm, 0cm, 0cm, 0cm}]{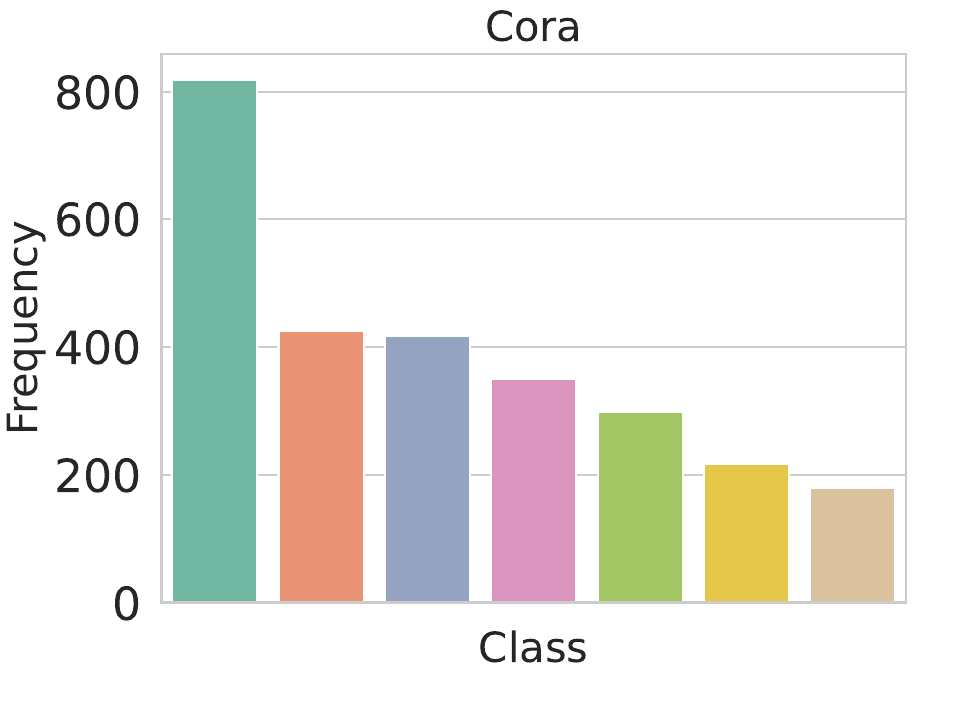}  
    \caption{Cora Label Frequency.}
\end{subfigure}
\begin{subfigure}{.32\textwidth}
    \centering
    \includegraphics[width=1\linewidth,trim={1cm, 0cm, 0cm, 0cm}]{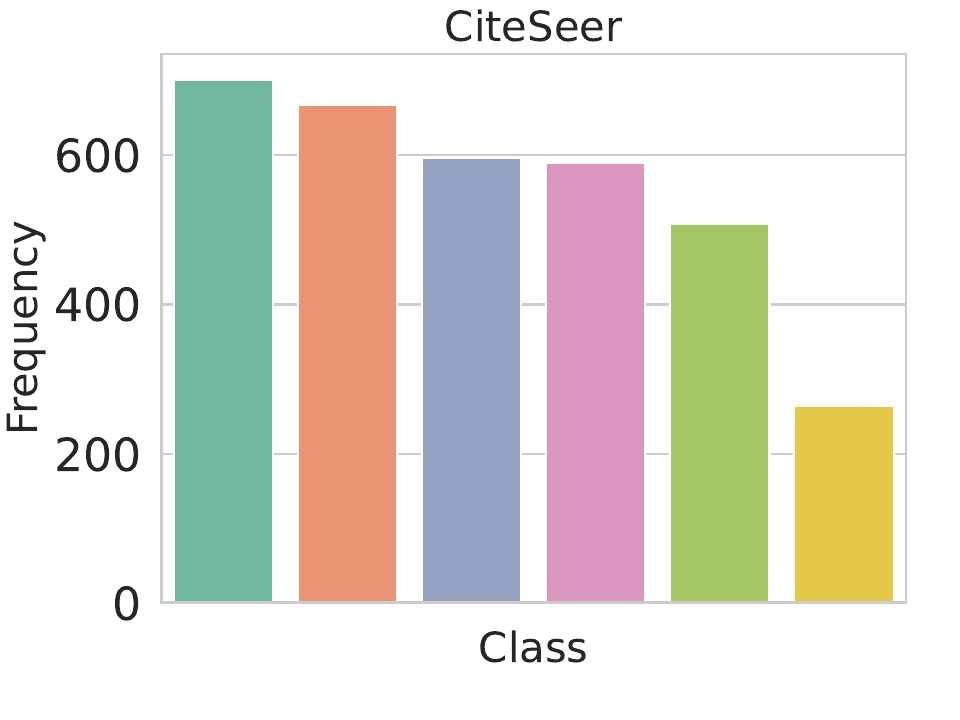}  
    \caption{CiteSeer Label Frequency.}
\end{subfigure}
\begin{subfigure}{.32\textwidth}
    \centering
    \includegraphics[width=1\linewidth,trim={1cm, 0cm, 0cm, 0cm}]{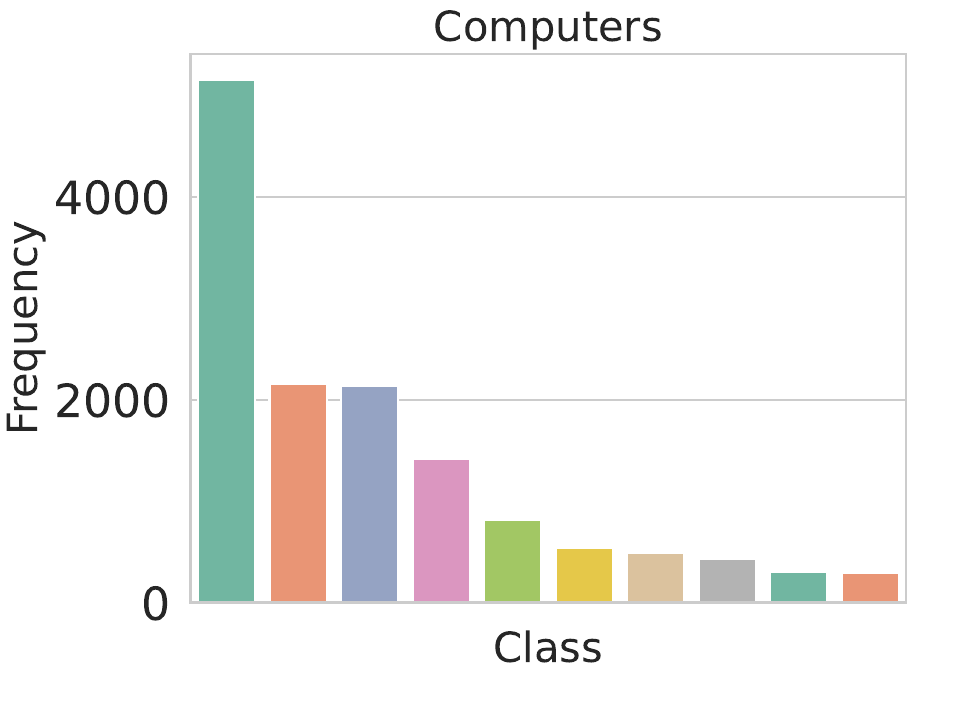}  
    \caption{Computers Label Frequency.}
\end{subfigure}

\vspace{0.5cm}

\begin{subfigure}{.32\textwidth}
    \centering
    \includegraphics[width=1\linewidth,trim={1cm, 0cm, 0cm, 0cm}]{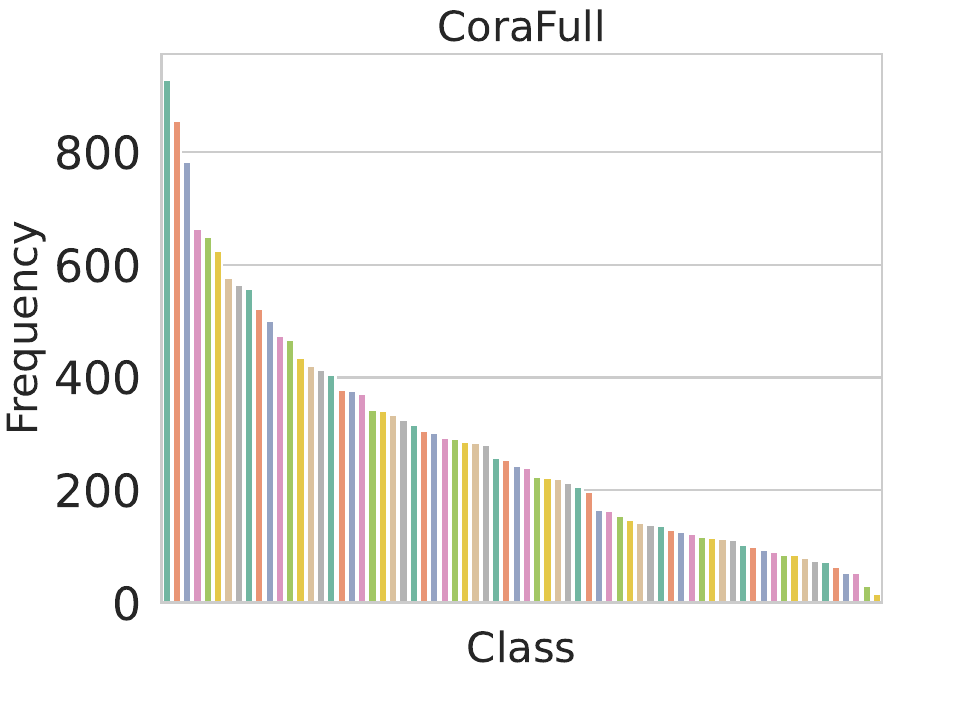}  
    \caption{CoraFull Label Frequency.}
\end{subfigure}
\begin{subfigure}{.32\textwidth}
    \centering
    \includegraphics[width=1\linewidth,trim={1cm, 0cm, 0cm, 0cm}]{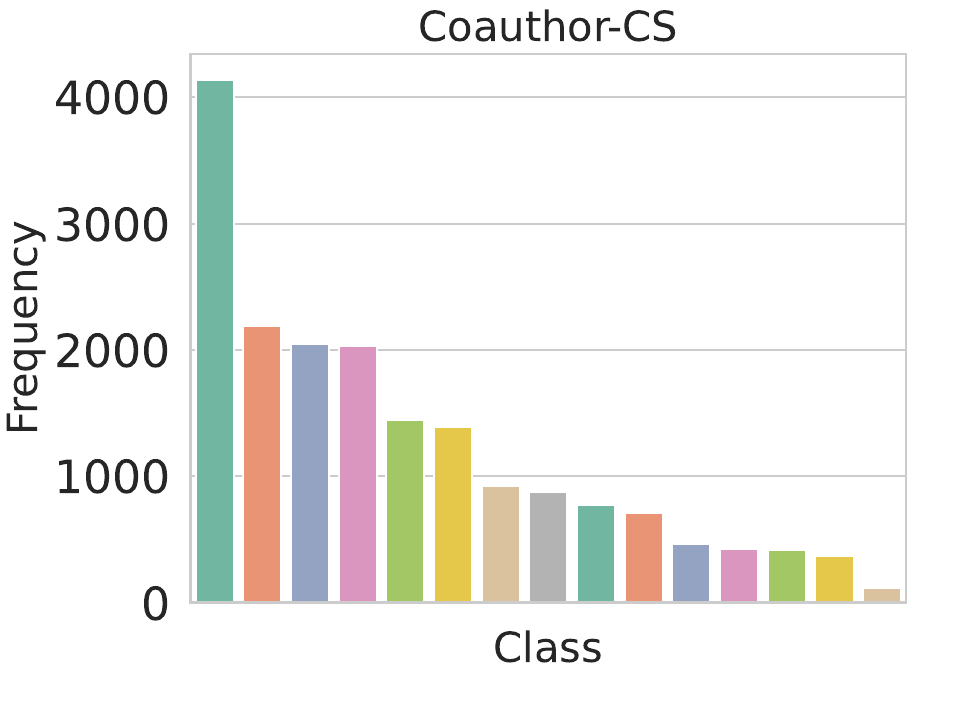}  
    \caption{Coauthor-CS Label Frequency.}
\end{subfigure}
\begin{subfigure}{.32\textwidth}
    \centering
    \includegraphics[width=1\linewidth,trim={1cm, 0cm, 0cm, 0cm}]{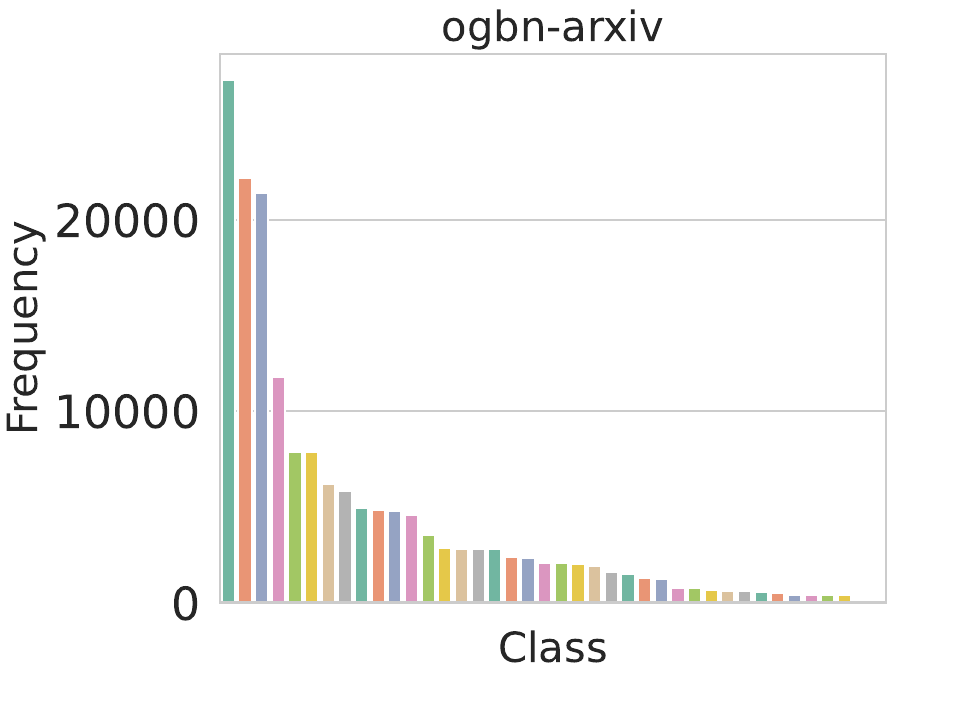}  
    \caption{ogbn-arxiv Label Frequency.}
\end{subfigure}
\caption{Label frequency distribution visualization of all datasets.}
\label{fig:data_distribution}
\end{figure}

We also present a histogram depicting the frequency of nodes within each category in Fig~\ref{fig:data_distribution}. As the figure illustrates, each dataset displays an imbalanced label distribution, with most exhibiting a long-tail distribution. This suggests that the complexity of each $N$-way $k$-shot meta-task will vary based on the sampled ways.

\vspace{-5pt}

\begin{table}[t]
\caption{Dataset Description and Meta-task Class Split Ratio.}
\setlength{\tabcolsep}{6pt}
\renewcommand{\arraystretch}{1.4}
\begin{tabular}{lccccccc}
\hline
                & \# of nodes & \# of edges & \# of feature & $C$  & $C_{train}$ & $C_{val}$ & $C_{test}$ \\ \hline
Cora~\cite{Cora-CiteSeer}            & 2,708        & 10,556       & 1,433          & 7  & 3            & 2          & 2           \\
CiteSeer~\cite{Cora-CiteSeer}        & 3,327        & 9,104        & 3,703          & 6  & 2            & 2          & 2           \\
Amazon-Computer~\cite{Coauthor-Amazon} & 13,752       & 491,722      & 767           & 10 & 4            & 3          & 3           \\
CoraFull~\cite{CoraFull}        & 19,793       & 126,842      & 8,710          & 70 & 38           & 15         & 15          \\
Coauthor-CS~\cite{Coauthor-Amazon}     & 18,333       & 163,788      & 6,805          & 15 & 5            & 5          & 5           \\
ogbn-arxiv~\cite{ogb}      & 169,343      & 1,166,243     & 128           & 40 & 20           & 10         & 10          \\ \hline
\end{tabular}
\label{tab:dataset}
\end{table}

\section{Reproducibility}
\label{app:repro}
We provide the source code of our method in this Github link:
\href{https://github.com/Haoliu-cola/COLA/}{https://github.com/Haoliu-cola/COLA}.

We report the average performance over 20 runs for each method and each task.  All models were tested on a single NVIDIA A100 80GB GPU. The experiments on all baselines follow the setting of previous work TLP~\cite{TLP}. In our method COLA, we use GCN with ReLU activation function as the graph encoder. For all three augmentation operations, we take the combination of randomly dropping edge and dropping feature and the dropping ratio for Lookup and Support Embeddings are set to the same value. We implement Grid Search in [0.1, 0.2, 0.3, 0.4, 0.5, 0.6, 0.7] to obtain the augmentation ratio for each dataset. During the meta-train phase, we take the average loss function over 20 pseudo-meta-tasks to obtain the final loss function. During the meta-test phase, we use the average performance of 100 meta-tasks and each meta-task includes 20 query samples for each way. Hyperparameters like learning rate, weight decay, and temperature parameter $\tau$ are presented in the YAML file of our provided codes.

\section{Additional Experiments}
\label{app:exp}

\subsection{Extra experiments of Section 3.1: ablation Study on GCL method with respect to nodes sampling and data augmentation}

In Section~\ref{sec:3.1}, we delved into the influence of classes from which nodes are sampled, and the use of data augmentation within the Graph Contrastive Learning (GCL) method GRACE~\cite{GRACE}. Here, we expand upon those results, providing further outcomes from additional datasets and tasks, with the aim of clarifying our preliminary findings. As the GRACE method employs all nodes in each optimization process, the experiment exceeds memory constraints with the ogbn-arxiv dataset; we present results for other datasets in the Figure.

Our conclusions are consistent across varied settings:

\begin{itemize}
    \item Leveraging all nodes within the graph to compute contrastive loss leads to superior performance when compared to using only a subset of nodes.
    \item Interestingly, there are instances where using nodes exclusively from $C_{test}$ results in comparable, or even superior, performance relative to the scenario where nodes from $C_{train} \cup C_{val}$ are used. Despite the latter containing significantly more nodes than the former, incorporating nodes from test sets proves crucial in avoiding overfitting.
    \item In the majority of cases, the application of data augmentation techniques enhances model performance, given that the core of contrastive learning is to ascertain invariant information across distinct views. However, in a few scenarios, the integration of data augmentation has an adverse effect on model performance. We conjecture one reason might be that when the model inherently lacks adequate classification capability, introducing data augmentation equates to adding noise, thereby harming the effective learning of representation. 
\end{itemize}

\begin{figure}
\centering
\begin{subfigure}{.44\textwidth}
    \centering
    \includegraphics[width=1\linewidth,trim={0cm, 0cm, 0cm, 0cm}]{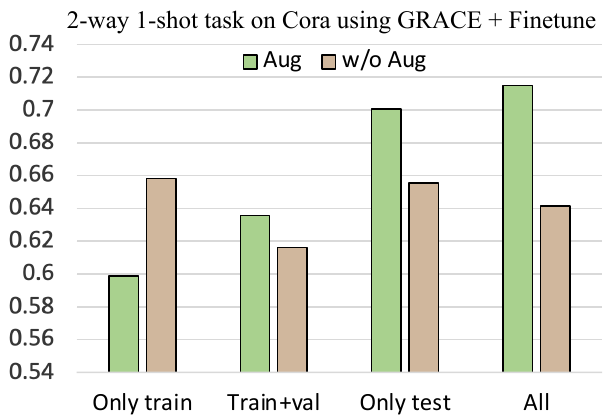}  
    \caption{Cora 2-way 1-shot task.}
\end{subfigure}
\begin{subfigure}{.44\textwidth}
    \centering
    \includegraphics[width=1\linewidth,trim={0cm, 0cm, 0cm, 0cm}]{figures/toy_cora25.pdf}  
    \caption{Cora 2-way 5-shot task.}
\end{subfigure}

\begin{subfigure}{.44\textwidth}
    \centering
    \includegraphics[width=1\linewidth,trim={0cm, 0cm, 0cm, 0cm}]{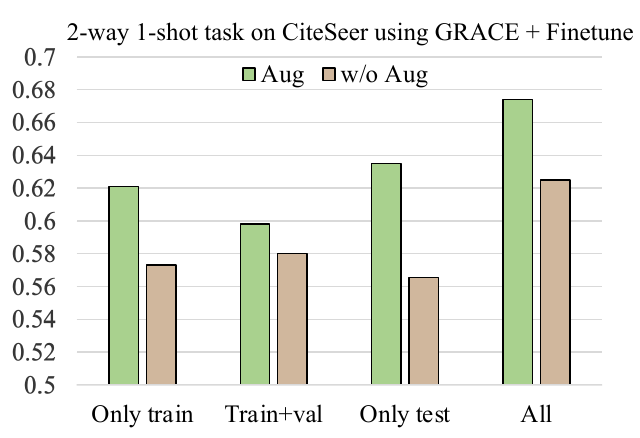}  
    \caption{CiteSeer 2-way 1-shot task.}
\end{subfigure}
\begin{subfigure}{.44\textwidth}
    \centering
    \includegraphics[width=1\linewidth,trim={0cm, 0cm, 0cm, 0cm}]{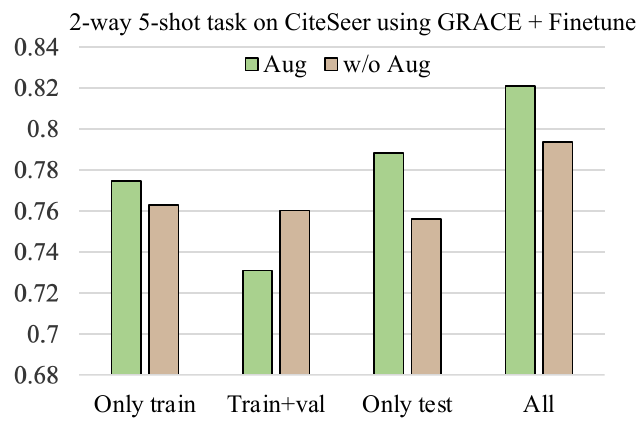}  
    \caption{CiteSeer 2-way 5-shot task.}
\end{subfigure}


\begin{subfigure}{.44\textwidth}
    \centering
    \includegraphics[width=1\linewidth,trim={0cm, 0cm, 0cm, 0cm}]{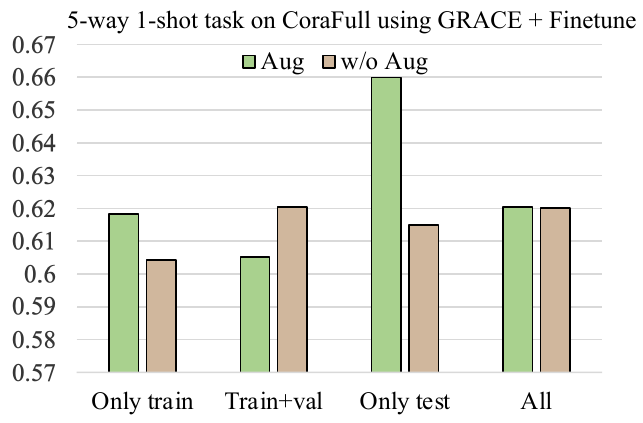}  
    \caption{CoraFull 5-way 1-shot task.}
\end{subfigure}
\begin{subfigure}{.44\textwidth}
    \centering
    \includegraphics[width=1\linewidth,trim={0cm, 0cm, 0cm, 0cm}]{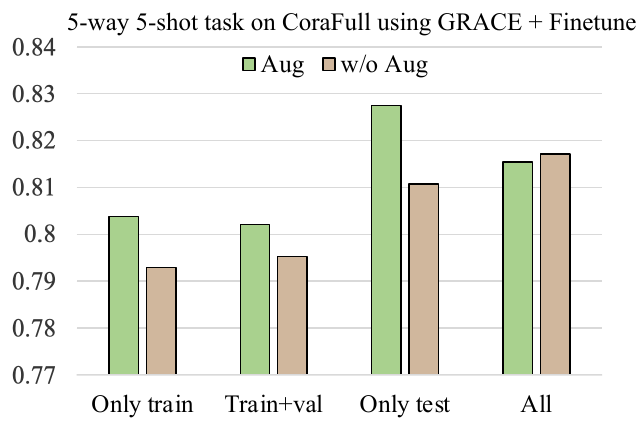}  
    \caption{CoraFull 5-way 5-shot task.}
\end{subfigure}

\begin{subfigure}{.44\textwidth}
    \centering
    \includegraphics[width=1\linewidth,trim={0cm, 0cm, 0cm, 0cm}]{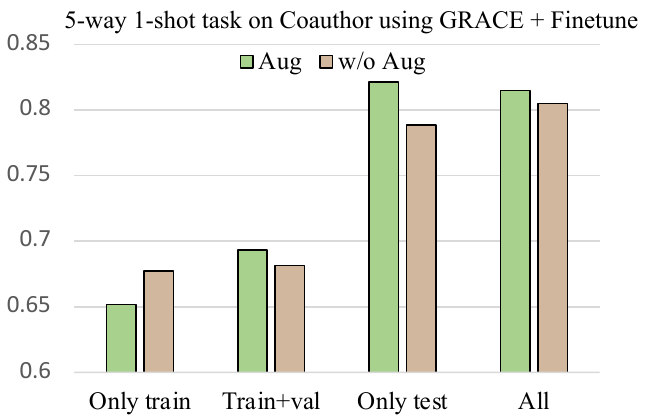}  
    \caption{Coauthor-CS 5-way 1-shot task.}
\end{subfigure}
\begin{subfigure}{.44\textwidth}
    \centering
    \includegraphics[width=1\linewidth,trim={0cm, 0cm, 0cm, 0cm}]{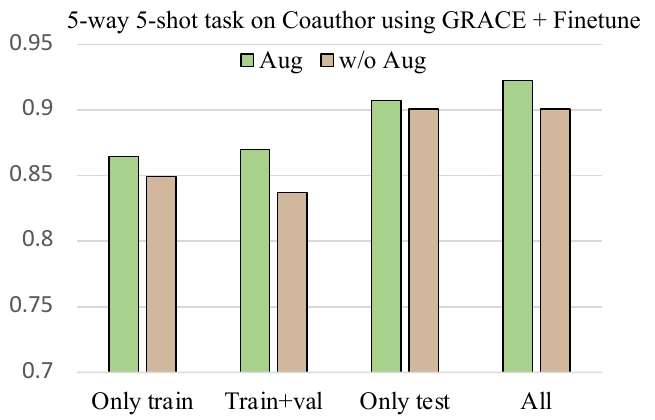}  
    \caption{Coauthor-CS 5-way 5-shot task.}
\end{subfigure}
\caption{Case study on GRACE+finetune framework.}
\label{app:toyfull}
\end{figure}

\subsection{Main results on Amazon-Computer, Coauthor-CS and ogbn-arxiv datasets}
We present our main results on Amazon-Computer, Coauthor-CS and ogbn-arxiv datasets in Table~\ref{tab:main2}. The results show that our method COLA still achieves the best performance in each task. The improvement is limited in some cases, e.g. 2-way 5-shot task of Amazon-Computer and the 5-way 5-shot task of Coauthor-CS, since the absolute accuracy is already close to 100\%. In ogbn-arxiv dataset, we can notice that almost all the contrastive-based GCL methods face the out-of-memory (OOM) issue, but COLA will not since we only involve a small subset of nodes to construct meta-tasks.
\begin{table}[t]
\small
\caption{Results on Amazon-Computer, Coauthor-CS and ogbn-arxiv datasets. (Top rows) Meta Learning. (Middle rows) Graph Contrastive Learning with fine-tuning. (Bottom row) COLA (our method). All scores are averaged over 20 runs. Evaluation metrics were scaled to 100 for readability purposes. In bold are methods with the best results for each task. In blue are methods with the best results in each group. $OOM$ indicates out of memory.}
\setlength{\tabcolsep}{4.4pt}
\renewcommand{\arraystretch}{1.3}
\begin{tabular}{@{}lcccccc}
\toprule
Dataset     & \multicolumn{2}{c}{Amazon-Computer}    & \multicolumn{2}{c}{Coauthor-CS} & \multicolumn{2}{c}{ogbn-arxiv} \\
\cmidrule(lr){2-3} \cmidrule(lr){4-5} \cmidrule(lr){6-7}
Task        & 2-way 1-shot & 2-way 5-shot & 5-way 1-shot  & 5-way 5-shot & 5-way 1-shot  & 5-way 5-shot \\
\midrule
\multicolumn{7}{c}{\cellcolor{pearDark!25}Meta learning}                                 \\
\midrule
MAML~\cite{finn2017model}                & 52.69 $\pm$ 2.23  & 59.19 $\pm$ 2.42  & 29.73 $\pm$ 1.54  & 43.78 $\pm$ 1.51   & 27.11 $\pm$ 1.49   & 28.83 $\pm$ 1.51      \\
ProtoNet~\cite{snell2017prototypical}    & 56.27 $\pm$ 2.54  & 63.11 $\pm$ 2.60  & 37.98 $\pm$ 1.69  & 51.10 $\pm$ 1.49   & 34.49 $\pm$ 1.72   & 46.21 $\pm$ 1.73  \\
Meta-GNN~\cite{zhou2019meta}             & 60.54 $\pm$ 2.79  & 68.36 $\pm$ 3.15  & 54.17 $\pm$ 2.02   & 67.24 $\pm$ 1.56  & 27.42 $\pm$ 1.96   & 32.08 $\pm$ 1.65  \\
GPN~\cite{GPN}                           & 57.59 $\pm$ 2.67  & 74.86 $\pm$ 2.27  & \cellcolor{pearDark!10}64.95 $\pm$ 1.43  & \cellcolor{pearDark!10}75.42 $\pm$ 1.56   & 36.23 $\pm$ 1.48   & 48.85 $\pm$ 1.60  \\
G-Meta~\cite{G-Meta}                     & 62.56 $\pm$ 3.11  & 71.47 $\pm$ 2.97  &  59.87 $\pm$ 2.35                & 73.16 $\pm$ 1.40                  & 26.45 $\pm$ 1.62   & 33.09 $\pm$ 1.65  \\
TENT~\cite{TENT}                         &\cellcolor{pearDark!10}77.74 $\pm$ 3.16  & \cellcolor{pearDark!10}86.06 $\pm$ 2.16  & 59.61 $\pm$ 1.87  & 74.84 $\pm$ 1.23   & \cellcolor{pearDark!10}47.55 $\pm$ 1.93   & \cellcolor{pearDark!10}61.98 $\pm$ 1.62  \\
\midrule
\multicolumn{7}{c}{\cellcolor{pearDark!25}Graph Contrastive Learning  $\pm$   Finetune}         \\
\midrule
BGRL~\cite{BGRL}         & 69.95 $\pm$ 3.15  & 83.99 $\pm$ 2.14  & 63.96 $\pm$ 2.19   & 89.53 $\pm$ 0.83 & 36.42 $\pm$ 1.70   & 53.63 $\pm$ 1.66  \\
MVGRL~\cite{MVGRL}       & 65.95 $\pm$ 2.76  & 85.22 $\pm$ 2.08  & 69.64 $\pm$ 2.15   & 89.27 $\pm$ 1.04 & $OOM$    & $OOM$   \\
MERIT~\cite{MERIT}       & 77.35 $\pm$ 1.87  &  \cellcolor{pearDark!10}95.19 $\pm$ 0.69  & 85.74 $\pm$ 1.61   & 96.40 $\pm$ 0.39   & $OOM$    & $OOM$   \\
GraphCL~\cite{GraphCL}   & 78.46 $\pm$ 3.05  & 93.53 $\pm$ 1.56  & 73.68 $\pm$ 2.49   & 89.74 $\pm$ 1.76  & $OOM$    &  $OOM$  \\
GRACE~\cite{GRACE}       & 75.83 $\pm$ 2.84  & 88.46 $\pm$ 2.12  & 81.50 $\pm$ 1.88   & 92.24 $\pm$ 0.73  & $OOM$    & $OOM$   \\
SUGRL~\cite{SUGRL}       & \cellcolor{pearDark!10}85.49 $\pm$ 2.07  & 95.13 $\pm$ 0.89  & \cellcolor{pearDark!10}92.47 $\pm$ 1.04   & \cellcolor{pearDark!10}96.78 $\pm$ 0.33  & \cellcolor{pearDark!10}57.46 $\pm$ 2.03   & \cellcolor{pearDark!10}76.03 $\pm$ 1.38  \\
\midrule
COLA (ours) & \textbf{87.52 $\pm$  1.78}   & \textbf{95.89 $\pm$  1.02}   & \textbf{93.23 $\pm$  1.27}    & \textbf{96.79 $\pm$ 0.68}   &\textbf{60.41 $\pm$  2.35}    & \textbf{77.40 $\pm$  2.09}  \\
\bottomrule

\end{tabular}
\label{tab:main2}
\end{table}

\section{Ablation Study Illustration}
\label{app:illustration}
We investigated three scenarios in Section~\ref{sec:ablation5.3.1} to verify the function of each component design. In this section, we first provide the illustration of these three scenarios for better understanding and then present the ablation study on other datasets. 

\begin{figure}
\centering
\begin{subfigure}{.9\textwidth}
    \centering
    \includegraphics[width=1\linewidth,trim={0cm, 0cm, 0cm, 0cm}]{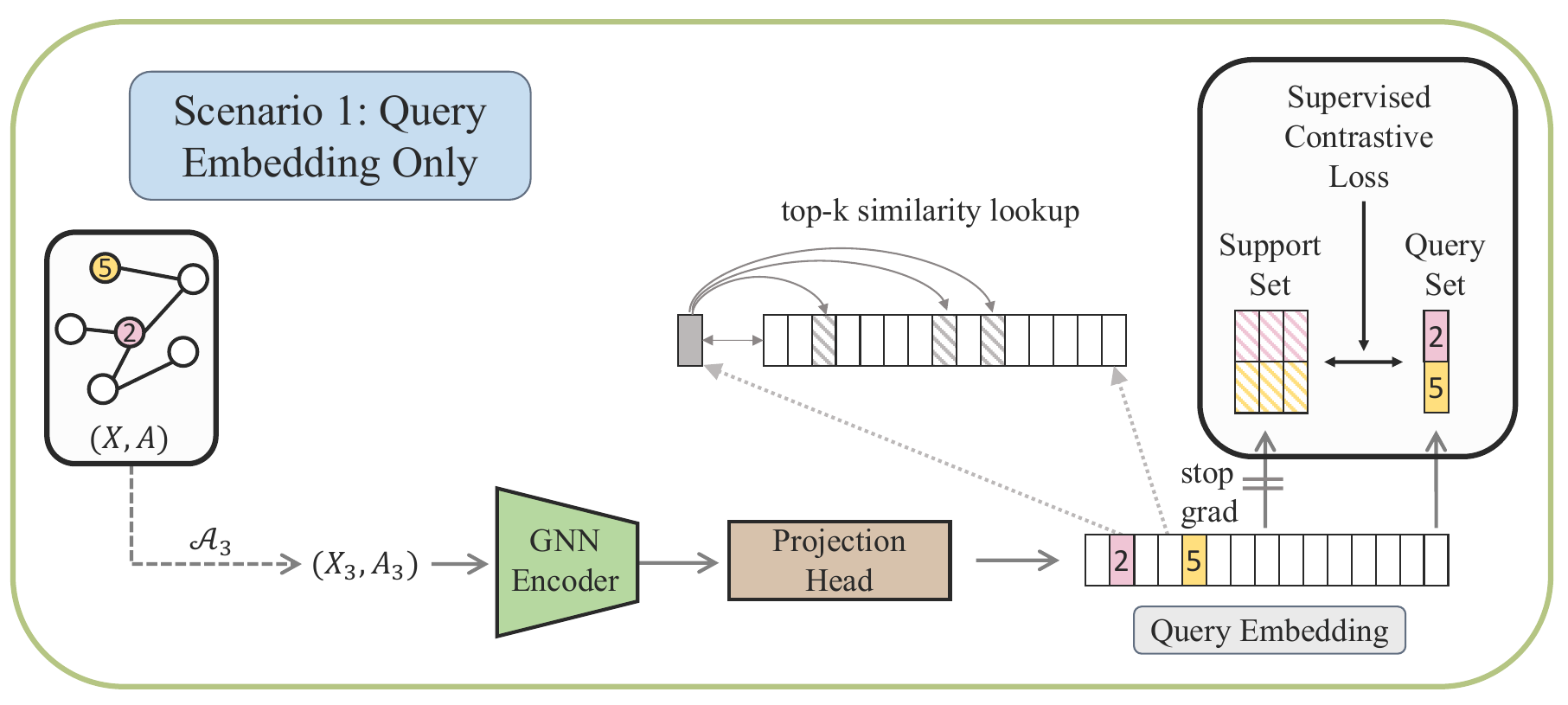}  
    \caption{Scenario 1: Only Query Embedding.}
    \label{fig:s1}
\end{subfigure}

\begin{subfigure}{.9\textwidth}
    \centering
    \includegraphics[width=1\linewidth,trim={0cm, 0cm, 0cm, 0cm}]{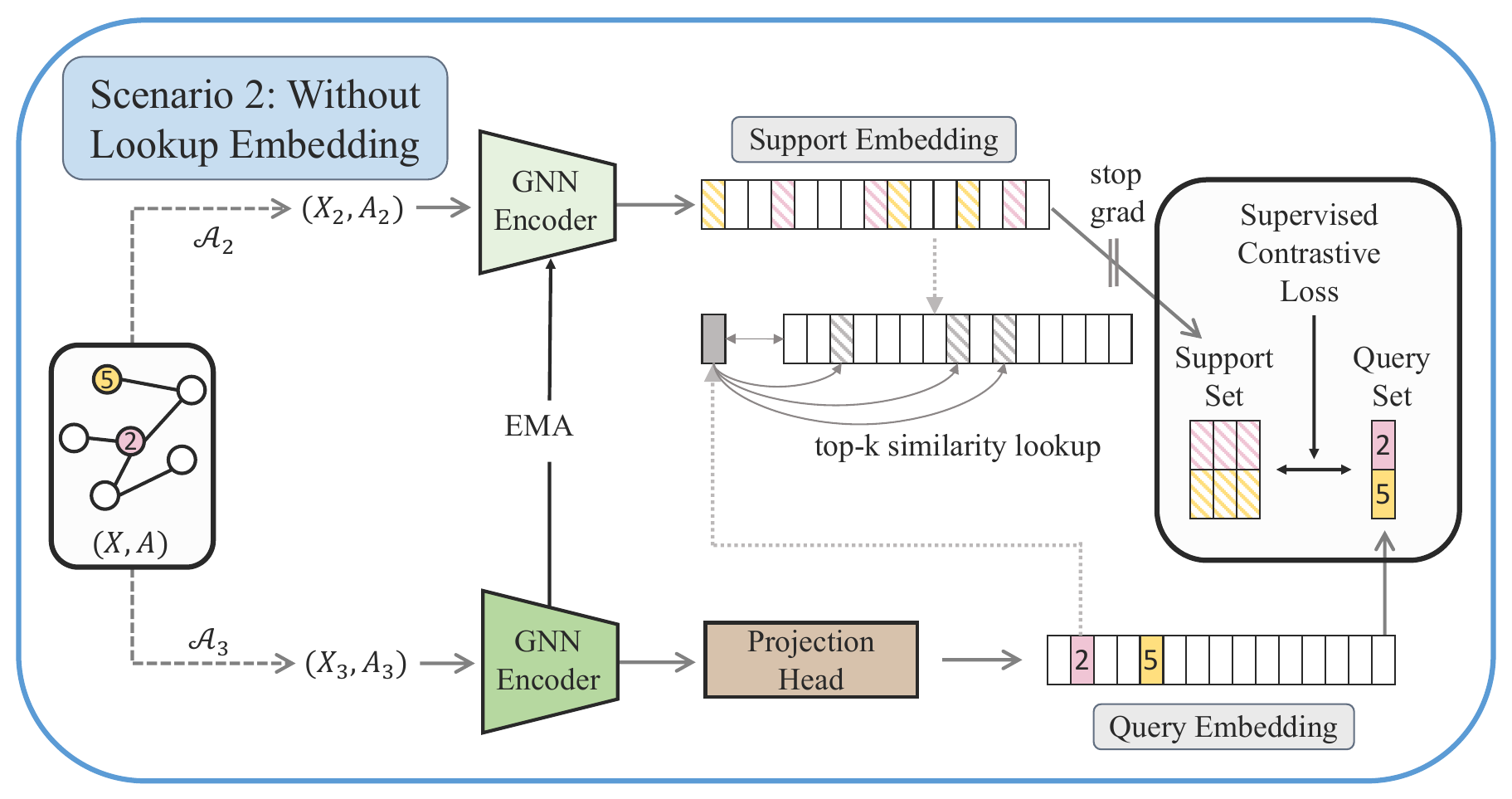}  
    \caption{Scenario 2: Without Lookup Embedding.}
    \label{fig:s2}
\end{subfigure}

\begin{subfigure}{.9\textwidth}
    \centering
    \includegraphics[width=1\linewidth,trim={0cm, 0cm, 0cm, 0cm}]{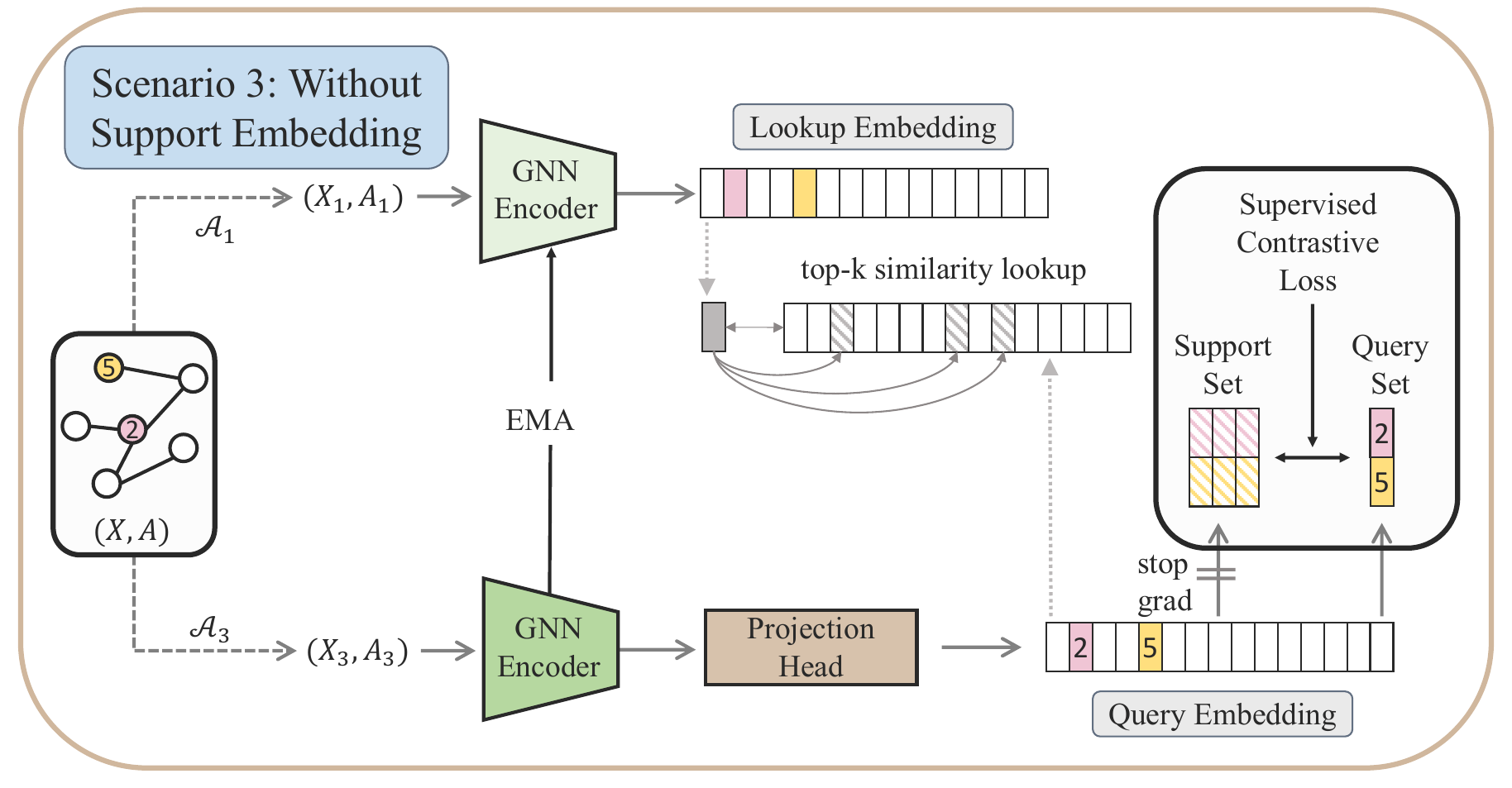}  
    \caption{Scenario 3: Without Support Embedding.}
    \label{fig:s3}
\end{subfigure}
\caption{Ablation Study Illustration.}
\label{app:ablationillustration}
\end{figure}

\begin{itemize}
    \item The first scenario is shown in Figure~\ref{fig:s1}, where we only have Query Embedding, thus both query and support sets are generated from Query Embedding itself.
    \item The second scenario is shown in Figure~\ref{fig:s2}. This scenario omits the Lookup Embedding, and the query embedding of the query node $v_i$ has to match with all node embeddings from Support Embedding.
    \item The third scenario is shown in Figure~\ref{fig:s3}. We discard the Support Embedding here, and the lookup embedding of the query node $v_i$ will match with all node embeddings from Query Embedding. Thus, both query and support sets are from Query Embedding.
\end{itemize}

We then provide the component analysis results on Amazon-Computer, CoraFull, Coauthor-CS and ogbn-arxiv datasets in Table~\ref{tabab2} and Table~\ref{tabab3}.

\begin{table}[t]
\small
\caption{Component Analysis of Query ($Q$), Support ($S$), Lookup ($L$) Embeddings on Amazon-Computer and CoraFull datasets. The first three rows control different components in meta-task construction. The last row is COLA's setting. In bold are the best results, and underlines are the second best ones. }
\setlength{\tabcolsep}{5pt}
\renewcommand{\arraystretch}{1.3}
\begin{tabular}{cccllllll}
\toprule
      &   &  & \multicolumn{3}{c}{Amazon-Computer}    & \multicolumn{3}{c}{CoraFull} \\
 \cmidrule(lr){4-6} \cmidrule(lr){7-9}
 $Q$& $S$& $L$& 2-way 1-shot  & 2-way 3-shot & 2-way 5-shot & 5-way 1-shot  & 5-way 3-shot & 5-way 5-shot \\
\midrule
\multicolumn{1}{c}{$\checkmark$} &                                                  & &   71.04 $\pm$ 2.07          & \underline{91.53 $\pm$ 2.26} & \underline{92.76 $\pm$ 2.34} & 58.96 $\pm$  1.99 & \underline{76.48 $\pm$ 2.74}& \underline{79.63 $\pm$ 2.39} \\
\multicolumn{1}{c}{$\checkmark$} & \multicolumn{1}{c}{$\checkmark$}                  & & 78.58 $\pm$ 2.61 & 85.87 $\pm$ 3.13 & 86.41 $\pm$ 2.45 & 64.62 $\pm$ 3.23 & 68.74 $\pm$ 2.10 & 71.43 $\pm$ 2.55 \\
\multicolumn{1}{c}{$\checkmark$} && \multicolumn{1}{c}{$\checkmark$}                & \underline{80.06 $\pm$ 1.78} & 88.28 $\pm$ 2.33 & 90.37 $\pm$ 2.89 & \underline{68.89  $\pm$ 2.09} & 75.17 $\pm$ 1.73 & 76.32 $\pm$ 2.76 \\

\multicolumn{1}{c}{\textcolor{pearDark!100}{$\checkmark$}} & \multicolumn{1}{c}{\textcolor{pearDark!100}{$\checkmark$}} & \multicolumn{1}{c}{\textcolor{pearDark!100}{$\checkmark$}} & \textbf{87.52 $\pm$  1.78}   & \textbf{93.08 $\pm$ 1.04}& \textbf{95.89 $\pm$  1.02}   & \textbf{74.36 $\pm$  2.37}   & \textbf{83.17 $\pm$  2.48} & \textbf{86.59 $\pm$  2.26}  \\
\bottomrule

\end{tabular}
\label{tabab2}
\end{table}
\begin{table}[t]
\small
\caption{Component Analysis of Query ($Q$), Support ($S$), Lookup ($L$) Embeddings on Coauthor-CS and ogbn-arxiv datasets. The first three rows control different components in meta-task construction. The last row is COLA's setting. In bold are the best results, and underlines are the second best ones. }
\setlength{\tabcolsep}{5pt}
\renewcommand{\arraystretch}{1.3}
\begin{tabular}{cccllllll}
\toprule
      &   &  & \multicolumn{3}{c}{Coauthor-CS}    & \multicolumn{3}{c}{ogbn-arxiv} \\
 \cmidrule(lr){4-6} \cmidrule(lr){7-9}
 $Q$& $S$& $L$& 5-way 1-shot  & 5-way 3-shot & 5-way 5-shot & 5-way 1-shot  & 5-way 3-shot & 5-way 5-shot \\
\midrule
\multicolumn{1}{c}{$\checkmark$} &                                                  & &   80.37 $\pm$ 2.86          & 90.45 $\pm$ 1.38 & 93.57 $\pm$ 1.19 & 30.17 $\pm$  2.36 & \underline{54.57 $\pm$ 2.04}& 58.94 $\pm$ 3.01 \\
\multicolumn{1}{c}{$\checkmark$} & \multicolumn{1}{c}{$\checkmark$}                  & & 82.21 $\pm$ 3.43 & 84.90 $\pm$ 2.59  & 90.46 $\pm$ 1.76 & 42.49 $\pm$ 1.97 & 45.27 $\pm$ 2.00 & 49.68 $\pm$ 2.36 \\
\multicolumn{1}{c}{$\checkmark$} && \multicolumn{1}{c}{$\checkmark$}                & \underline{88.75 $\pm$ 1.96} & \underline{92.39 $\pm$ 1.73} & \underline{94.53 $\pm$ 1.87} & \underline{50.88  $\pm$ 2.73} & 53.96 $\pm$ 3.25 & \underline{61.05 $\pm$ 2.84} \\

\multicolumn{1}{c}{\textcolor{pearDark!100}{$\checkmark$}} & \multicolumn{1}{c}{\textcolor{pearDark!100}{$\checkmark$}} & \multicolumn{1}{c}{\textcolor{pearDark!100}{$\checkmark$}} & \textbf{93.23 $\pm$  2.17}   & \textbf{96.42 $\pm$ 1.25}& \textbf{96.79 $\pm$  0.68}   & \textbf{60.41 $\pm$  2.35}   & \textbf{69.74 $\pm$  2.28} & \textbf{77.40 $\pm$  2.09}  \\
\bottomrule

\end{tabular}
\label{tabab3}
\end{table}

\section{Limitation}
\label{app:limitation}

One limitation of COLA is the computational cost due to the way of meta-task construction, involving cosine similarity computation between the query node and all graph nodes, followed by a sort to obtain the top-$k$ nodes. Assuming a graph with $|\mathcal{V}|$ nodes and $|\mathcal{E}|$ edges, and node embeddings with dimension $d$. Given $t$ $n$-way $k$-shot meta-tasks per training batch, the cosine similarity's time complexity is  $O(|\mathcal{V}|dtn)$, that of sorting operation is $O(|\mathcal{V}|\sqrt{k}t)$, that of MPGNN is $O(|\mathcal{E}|)$. Thus, the time complexity of our method is $O(|\mathcal{V}|dtn+|\mathcal{E}|)$. Excluding the GNN, with $d,t,n$ all being constants, the complexity remains linear with respect to the number of nodes. We illustrate the convergence time (in seconds) across different datasets in Table~\ref{tabcomplexity}. Although our convergence time is relatively longer than most baselines, this marginal increase is justifiable given the notable performance improvement.


\begin{table}[t]
\centering
\caption{Convergence time comparison (in seconds) on a single NVIDIA A100 80GB GPU. }
\renewcommand{\arraystretch}{1.3}
\begin{tabular}{@{}ccccc@{}}
\toprule
         & Cora 2-way 1-shot & Cora 2-way 5-shot  & CoraFull 5-way 1-shot & CoraFull 5-way 5-shot \\ \midrule
MAML     & 13.15    & 10.42    & 22.71        & 18.16        \\
ProtoNet & 17.40    & 16.83    & 31.39        & 19.38        \\
Meta-GNN & 26.33    & 25.03    & 92.99        & 83.32        \\
GPN      & 13.3     & 10.67    & 34.43        & 53.04        \\
G-Meta   & 46.62    & 191.82   & 196.01       & 662.54       \\
TENT     & 64.46    & 43.90    & 58.12        & 58.92        \\
BGRL     & 13.89    & 12.98    & 36.58        & 41.41        \\
MVGRL    & 98.23    & 110.56   & 654.79       & 707.63       \\
MERIT    & 955.60   & 1461.97  & 6240.12      & 8341.16      \\
GraphCL  & 62.37    & 70.78    & 450.17       & 502.64       \\
GRACE    & 8.38     & 6.80     & 74.42        & 41.53        \\
SUGRL    & 25.05    & 16.07    & 542.86       & 428.57       \\
COLA     & 83.43    & 103.64   & 619.65       & 817.43       \\ \bottomrule
\end{tabular}
\label{tabcomplexity}
\end{table}

Another limitation is that we do not explore much on different choices of the loss function and take the supervised contrastive loss~\cite{khosla2020supervised} since in this work we focus more on the method to construct meta-tasks without labels. Future work could explore more on this aspect based on the meta-task construction.

\end{document}